%% file: chil-sample.tex
\documentclass[pmlr,twocolumn,10pt]{jmlr} % W&CP article

\usepackage{booktabs}
\usepackage{siunitx}

\usepackage[switch]{lineno}

\usepackage{algorithm}
\usepackage{algorithm2e}
\usepackage{microtype}
\usepackage{graphicx}
\usepackage{booktabs}
\usepackage{xspace}
\usepackage{multirow}
\usepackage{wrapfig}
\usepackage{enumitem}
\usepackage{titletoc}
\usepackage[defaultcolor=magenta]{changes}
\newcommand{\eg}{\emph{e.g.,}\xspace}

\newcommand{\ie}{\emph{i.e.,}\xspace}
\newcommand{\mine}{\textsc{H-AdminSim}\xspace}
\usepackage{caption}
\usepackage{color}
\usepackage[most]{tcolorbox}
\usepackage{bbding}
\usepackage{pifont}
\usepackage{makecell}
\usepackage{xurl}
\usepackage{colortbl}
\usepackage{subcaption,siunitx,booktabs}
\usepackage[defaultcolor=magenta]{changes}

\theorembodyfont{\upshape}
\theoremheaderfont{\scshape}
\theorempostheader{:}
\theoremsep{\newline}

\jmlrvolume{333}
\jmlryear{2026}
\jmlrsubmitted{LEAVE UNSET}
\jmlrpublished{LEAVE UNSET}
\jmlrworkshop{Conference on Health, Inference, and Learning (CHIL) 2026} % W&CP title
\title[\mine]{\mine: A Multi-Agent Simulator for Realistic Hospital Administrative Workflows with FHIR Integration}

\author{%
}

\author{%
\Name{Jun-Min Lee} \Email{ljm565@kaist.ac.kr}\\
\addr KAIST, Republic of Korea
\AND
\Name{Meong Hi Son} \Email{meonghi.son@samsung.com}\\
\addr Samsung Medical Center, Republic of Korea
\AND
\Name{Edward Choi} \Email{edwardchoi@kaist.ac.kr}\\
\addr KAIST, Republic of Korea
}

\begin{document}

\maketitle

\begin{abstract}
Hospital administration departments handle a wide range of operational tasks and, in large hospitals, process over 10,000 requests per day, driving growing interest in LLM-based automation. 
However, prior work has focused primarily on patient--physician interactions or isolated administrative subtasks, failing to capture the complexity of real administrative workflows.
To address this gap, we propose \mine, a comprehensive simulation framework that combines realistic data generation with multi-agent–based simulation of hospital administrative workflows.
These tasks are quantitatively evaluated using detailed rubrics, enabling systematic comparison of LLMs. 
Through FHIR integration, \mine provides a unified and interoperable environment for testing administrative workflows across heterogeneous hospital settings, serving as a standardized testbed for assessing the feasibility and performance of LLM-driven administrative automation.
\end{abstract}

\paragraph*{Data and Code Availability}
All data used in this study were synthetically generated.
The code is publicly available on our project page.\footnote{\url{https://ljm565.github.io/h_adminsim_project/}}

\paragraph*{Institutional Review Board (IRB)}
This research does not require IRB approval.

\section{Introduction}
\label{sec:Introduction}
% Backgrounds
Hospital administration serves as the patient’s first point of contact and handles a wide range of administrative tasks, including medication, insurance, and diagnostic inquiries, as well as department assignment and outpatient scheduling~\citep{Litchfield523,Bhati2023-sa}.
Larger hospitals often process more than 10,000 patient requests per day~\citep{Lin2020-py,wan_outpatient_2024}.
Although larger hospitals employ more administrative staff, prior studies have reported that several administrative staff members lack formal medical training~\citep{oh2019relationship,Lin2025}.
Unlike returning patients, individuals visiting a hospital for the first time with a given health condition often lack prior medical referrals, making accurate department assignment challenging even for experienced administrative staff~\citep{Kajiwara2017-ep} and potentially leading to delays in appropriate care.
Moreover, appointment scheduling requires administrative staff to coordinate physician availability and patient preferences within the hospital system while handling ongoing requests from patients with existing appointments, further increasing administrative complexity~\citep{gupta2008appointment}.

Given these operational complexities, large language model (LLM)-based agents offer a promising avenue for automating or assisting hospital administrative workflows. 
However, their safe deployment requires prior validation in simulation environments that closely replicate real-world processes. 
Despite this need, existing healthcare simulation research has largely focused on patient--physician interactions, either through dialogue-based role-play for clinical education and training~\citep{wang-etal-2024-patient,li2024leveraginglargelanguagemodel} or simulated clinical dialogues for evaluating LLM behavior~\citep{fan-etal-2025-ai,almansoori2025selfevolvingmultiagentsimulationsrealistic}.
Even studies incorporating administrative components have primarily addressed physician-facing scenarios~\citep{yan2024clinicallab} or narrow subtasks, often overlooking time-dependent scheduling workflows that are central to hospital administration~\citep{PIORS}. 
Furthermore, although patient and physician characteristics differ markedly across hospital levels~\citep{wiki:xxx,hospital-level}, few studies have comprehensively modeled these differences. 

To address these gaps, we propose \mine, a framework for realistic dialogue-based simulation of outpatient hospital administrative workflows with three core components:
(1) simulation environments for hospitals of varying scales, from primary to tertiary care;
(2) dialogue-based patient intake simulations that capture diverse patient backgrounds, modeled using 194 disease--symptom pairs and information on whether the patient has prior referrals; and
(3) dialogue-based scheduling simulations for new appointments, rescheduling, and cancellation that reflect individual scheduling preferences.
Additionally, our simulation environment follows the Fast Healthcare Interoperability Resources (FHIR) standard~\citep{6627810}, enabling interoperability across heterogeneous institutions. 
While previous FHIR-based research has largely focused on API-based benchmarking~\citep{jiang2025medagentbenchrealisticvirtualehr} or data transformation~\citep{10.1093/jamia/ocad002,CARBONARO2025111051}, to the best of our knowledge, \mine is the first to leverage FHIR for dialogue-based hospital administration simulation~\citep{info:doi/10.2196/21929}.
Overall, \mine provides a standardized environment for systematic LLM evaluation in hospital administration and a simulation-based platform for assessing the feasibility of LLM-driven workflow automation.

\section{Related Works}
\label{appendix:related works}
Most existing LLM-based healthcare simulation studies focus on clinical, physician-centered settings, where models are evaluated primarily on diagnostic accuracy and clinical reasoning through patient--physician interactions.
Several works propose multi-agent frameworks in which multiple physician LLMs collaborate to improve clinical decision-making, typically benchmarked on medical question--answering datasets~\citep{tang2024medagents,kim2024mdagents}.
In addition, task-level workflow simulations usually assign LLMs the role of physicians and evaluate diagnostic and clinical reasoning in dialogue-based clinical scenarios~\citep{schmidgall2024agentclinic,liu2024medpmcmedicalpersonalizedmultimodal,liao2024automaticinteractiveevaluationlarge,fan-etal-2025-ai}.
MedAgentSim~\citep{almansoori2025selfevolvingmultiagentsimulationsrealistic} further introduces a self-improvement mechanism that iteratively enhances physician performance within such clinical simulations.
LLM-based agent simulations have also been explored as educational tools, using role-playing with simulated patients to support clinical training and communication skills~\citep{li2024leveraginglargelanguagemodel,wang-etal-2024-patient,wei2024medco,hicke2025medsimai}.

Among these simulation studies, only a few introduce limited administrative components. 
For example, Agent Hospital~\citep{li2024agent} and ClinicalLab~\citep{yan2024clinicallab} include a department guidance stage, where nurse-like agents assign patients to departments based on a single chief complaint or brief symptom descriptions. 
However, these components do not capture naturalistic intake workflows: they omit dialogue-based elicitation of patient symptoms and collection of patient demographics, and instead remain centered on patient--physician interactions and clinical reasoning. 
PIORS~\citep{PIORS} proposes an outpatient administrative simulation focused on reception nurses, but its scope is restricted to conversational intake and department guidance, without modeling temporal constraints or scheduling workflows that are central to real-world operations. 
In contrast, our work introduces a hospital administration-centric simulation that integrates dialogue-based intake with temporally constrained scheduling workflows for realistic evaluation of outpatient administration.

% Among these simulation studies, only a few introduce limited administrative components.
% For example, Agent Hospital~\citep{li2024agent} and ClinicalLab~\citep{yan2024clinicallab} include a department guidance stage, where nurse-like agents assign patients to departments based on a single chief complaint or brief symptom descriptions.
% However, these components do not capture naturalistic intake workflows: they omit dialogue-based elicitation of patient symptoms and collection of patient demographics, and instead remain centered on patient--physician interactions and clinical reasoning.
% PIORS~\citep{PIORS} focuses on reception nurses for conversational intake and department guidance, but does not model temporal constraints or scheduling workflows.
% In contrast, our work provides a hospital administration-centric simulation that integrates dialogue-based intake with temporally constrained scheduling for realistic outpatient evaluation.

\section{Task Definition}
This study focuses on first-visit outpatient workflows as a core subset of hospital administrative processes. 
A first-visit encounter is a canonical setting in which administrative staff directly elicit patient information and make foundational decisions under uncertainty, such as department assignment and appointment scheduling. 
In contrast, follow-up visits build on previous encounters and involve less extensive administrative intake, though they may require more complex scheduling for lab tests, imaging, or other procedures.
Given this distinction, we restrict our scope to first-visit workflows, which serve as the entry point to outpatient care and the main stage for administrative decision-making.

\noindent\textbf{Patient Intake}. 
During patient intake, administrative staff collect demographic and medical information through dialogue-based interactions, including their chief complaint and, in higher-level hospitals, prior medical referrals.
This information forms the basis for department assignment and appointment scheduling.

\noindent\textbf{Appointment Scheduling}. 
Once the patient’s department is determined, administrative staff elicit the patient's scheduling preferences through dialogue and make an appointment accordingly.
In higher-level hospitals, physician capacity---the number of outpatients a physician sees per hour---varies by department and individual consultation style and is managed to prevent physician overburdening.
Staff also handle rescheduling and cancellation requests from patients with existing appointments.

\noindent\textbf{FHIR Integration}. 
After completing these tasks, staff record patient demographics and appointment information in the hospital information system (HIS).
However, HIS schemas differ across hospitals, leading to inconsistencies in data representation that hinder interoperability~\citep{MARFOGLIA2025109745}.
To mitigate this, many countries have adopted the FHIR standard for representing and exchanging medical data~\citep{10.3389/fdgth.2024.1480600}.
In this study, we instantiate core FHIR resources to represent these administrative workflows.

\section{\mine: Hospital Outpatient Administration Simulation}
\begin{figure*}[t]
    \includegraphics[width=1\textwidth]{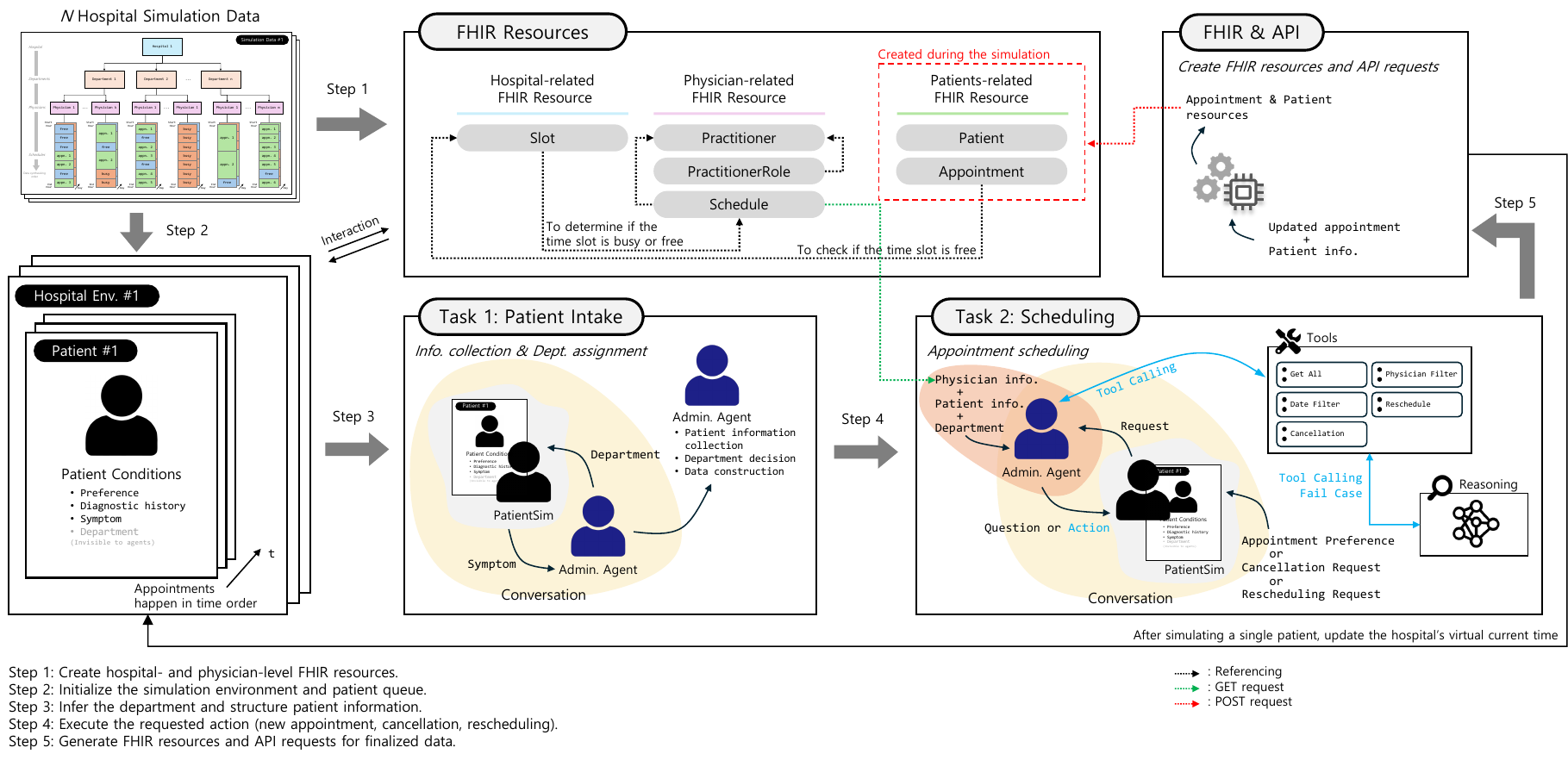}
    \caption{Diagram of the hospital administration simulation. Synthesized hospital data populate the hospital information system (HIS) with physician and hospital information using FHIR. The framework then simulates patient intake and appointment scheduling, uploading the resulting patient and appointment records to the HIS while keeping physician schedules updated via real-time FHIR communication. The hospital simulation data shown in the upper-left are shown in Appendix~\ref{appendix:data synthesis details}.}

    \label{fig:simulation}
\end{figure*}
We present \mine, a simulation framework for first-visit outpatient administration in hospitals.
The framework covers synthetic data generation, multi-agent simulation of administrative workflows, and integration with HIS via the FHIR standard.

\subsection{Simulation Data Synthesis}
\mine generates synthetic data following a hierarchical generation process with three stages: hospital, physician, and patient.
Further details are provided in Appendix~\ref{appendix:data synthesis details}.

\subsubsection{Hospital Configurations}
Hospital configurations specify the time system as well as the hospital’s department information.
The time system defines the hospital’s operating hours (\eg 10:00--17:00), time unit, and simulation period.

\noindent\textbf{Time Unit}. 
The time unit ($\tau$) represents the minimum temporal interval in hospital operation modeling and serves as the foundation for all time-dependent features, including physician schedules and appointments.
All time-related FHIR resources reference time slots generated based on $\tau$.
For example, an 8-hour hospital day with $\tau = 0.25$ hours (15 minutes) is divided into 32 time slots, with appointments and physician schedules managed accordingly.

\noindent\textbf{Departments}. 
This study focuses on nine internal medicine specialties that are widely recognized in clinical practice: gastroenterology, cardiology, pulmonology, endocrinology/metabolism, nephrology, hematology/oncology, allergy, infectious diseases, and rheumatology.\footnote{References are provided in Appendix~\ref{appendix:hospital_data}}
Because internal medicine exhibits substantial symptom overlap, department assignment becomes particularly challenging and error-prone, underscoring the need for further study on assignment accuracy~\citep{graber2005diagnostic,jain2014workplace}.
In this stage, users can specify a range for the number of physicians per department, and the actual number is randomly determined within that range.

\subsubsection{Physician Configurations}
Given the department-level physician counts defined in the hospital configuration, this stage generates physician profiles with randomly assigned demographics.

\noindent\textbf{Working Days}. 
In smaller hospitals, physicians typically provide outpatient services on all weekdays, whereas those in higher-level hospitals follow variable schedules, offering outpatient clinics only when they are not assigned to other duties such as inpatient care or research.
To reflect these differences across hospital levels, the data generation process allows users to specify a range for the number of outpatient working days per physician within the simulation period.

\noindent\textbf{Capacity}. 
Physician capacity denotes the number of outpatients a physician can see per hour~\citep{capacity}, with a theoretical upper bound of $1/\tau$, though practical capacity varies with each physician’s average consultation time.
During data synthesis, users specify a capacity range, and each physician is randomly assigned a capacity within this range, capped by this upper bound.

\noindent\textbf{Schedule}. 
For each physician, two types of time slots---busy schedules and outpatient appointments---are prefilled on outpatient working days during data synthesis.
The number of slots per appointment is determined by the physician’s capacity---for example, with $\tau = 0.25$ hours and a capacity of 2, each appointment occupies two consecutive slots (0.5 hours).
Users can specify the daily proportions of busy schedules and appointments, enabling random generation of physician schedules within the defined bounds.

\subsubsection{Patient Configurations}
\label{sec:patient conditions}
After prefilling virtual appointments for physicians, patient profiles are created to match the number of appointments, with demographic and condition-related attributes randomly assigned.

\noindent\textbf{Scheduling Preferences}.
Patient appointment scheduling preferences are categorized into three types that capture common patterns in real-world hospital administration: (1) earliest possible consultation ($asap$); (2) earliest consultation after a specified date ($date$); and (3) a preferred physician ($physician$)~\citep{douglas2005rheumatology,patrick2008dynamic,AGRAWAL20231318}.
In this study, these three preference types are combined into primary and secondary preferences, yielding six possible preference combinations per patient and reflecting the diverse ways in which preferences are expressed in administrative settings.
Secondary preferences are probabilistically activated during the simulation.

\noindent\textbf{Disease Profile}. 
To construct patient profiles for simulation, each patient is assigned diseases and associated symptoms.
Gold-standard labels specifying the correct target department for each disease are required to evaluate agent performance in department assignment.
Symptom data for each disease were collected from the National Health Service (NHS) health encyclopedia,\footnote{\href{https://www.nhsinform.scot/illnesses-and-conditions/a-to-z/}{NHS Inform Health Encyclopedia Scotland}} and department labels were created based on medical reference materials.
From the 427 diseases listed by the NHS, 194 related to internal medicine were selected to build disease profiles that include department labels and symptom information across the nine specialties. 
Details of the labeling process are provided in Appendix~\ref{appendix:disease labeling}.

\noindent\textbf{Diagnostic History}. 
Patients are categorized as either (1) those without a prior diagnosis ($without\_history$) or (2) those with a prior diagnosis ($with\_history$).
In primary and secondary care, patients generally visit without prior diagnoses, whereas in tertiary care many patients visit after receiving diagnoses at lower-level institutions and may prefer physicians specializing in their diagnosed condition~\citep{wiki:xxx,hospital-level}.
These diagnostic statuses are generated during data synthesis using predefined probabilities to simulate various hospital scenarios.

\subsection{FHIR-Integrated Hospital Environment}
\textbf{FHIR Resources}. 
After generating the simulation data, essential hospital information is uploaded to the HIS via FHIR to construct a realistic environment.
Each physician is represented by \textit{Practitioner} and \textit{PractitionerRole} resources, and their working days are encoded in a \textit{Schedule} resource.
\textit{Slot} resources, segmented by the time unit ($\tau$), are linked to each \textit{Schedule} and indicate outpatient availability as \textit{busy} or \textit{free}, forming the reference timetable for agents.
Patient information and scheduling outcomes are recorded in \textit{Patient} and \textit{Appointment} resources during task execution.
Further details of the FHIR resources are provided in Appendix~\ref{appendix:fhir resource}.

\noindent\textbf{Time Flow}. 
Hospital administrative tasks are inherently time-dependent, making temporal flow a key component of the simulation.
For instance, appointments cannot be scheduled in past time slots, and ongoing or completed appointments cannot be canceled.
To capture such constraints, the simulation explicitly models temporal progression using a virtual current time, which can be initialized up to $n$ days before the start of the simulation period.
As time advances, appointment statuses are updated (\textit{scheduled}, \textit{in-progress}, \textit{completed}), and rescheduling or cancellation is allowed only for \textit{scheduled} appointments.

\subsection{Simulation of Administration Tasks}
As shown in Figure~\ref{fig:simulation}, \mine simulates hospital administrative workflows to emulate real-world outpatient administration, enabling systematic evaluation of administrative agents.
The simulation environment comprises a patient agent and a staff agent responsible for administrative tasks.

\subsubsection{Patient Intake} 
During the patient intake simulation, patient and administrative staff agents engage in a dialogue.
The staff agent follows a sequence that begins with greeting, proceeds to collecting demographic information and asking about symptoms and prior diagnoses, and concludes with a department recommendation.
Throughout the interaction, the staff agent maintains polite and empathetic communication while minimizing technical jargon~\citep{Kwame2021_nurse,Sharkiya2023_nursequality}.
The patient agent’s persona is generated by integrating the synthesized patient profile---which includes scheduling preferences, prior diagnosis status, and disease--symptom information---with \textsc{PatientSim} attributes~\citep{kyung2025patientsimpersonadrivensimulatorrealistic} such as personality, history recall level, confusion level, and language proficiency.
Patients without prior diagnoses describe only their symptoms without referencing specific diseases.
At the end of the dialogue, the staff agent extracts and structures the patient’s demographic information and department recommendation for downstream tasks, and these outputs are evaluated using rubrics on data format correctness, simulation completeness, department appropriateness, and information extraction quality.
Appendix~\ref{appendix:prompts:intake} provides details on \textsc{PatientSim} attributes and prompts, and Appendix~\ref{appendix:simulation details:simulation_dialogues} presents example simulation dialogues.

\subsubsection{Appointment Scheduling}
After patient intake, the staff agent elicits the patient's scheduling preferences through dialogue and proposes an appointment based on physician availability in the assigned department and the current simulation time.
When necessary, the agent asks clarifying questions until the preferences are fully identified, then queries the FHIR-based hospital environment to retrieve available outpatient slots.
The patient agent first states a primary preference and may reject the proposed schedule with a 30\% probability, in which case a secondary preference is provided.
This design yields nine scheduling dialogue types, formed from three primary preference types and their six possible primary--secondary combinations.
The staff agent identifies the patient’s preferences from the dialogue and, when it determines that a scheduling tool is needed, invokes the appropriate tool within the assigned department:
(1) the earliest available slot across all physicians ($asap$); 
(2) the earliest slot after a specified date ($date$); or 
(3) the earliest slot for a preferred physician ($physician$).
Tool-based scheduling (\textit{Scheduling (T)}) requires tools that cover diverse patient preferences but yields stable performance.
On the other hand, reasoning-based scheduling (\textit{Scheduling (R)}) is prompt-based and easy to implement, yet performs substantially worse (up to 75.9 percentage points lower; Table~\ref{tab:simulation results}). 
Consequently, our framework uses tool-based scheduling by default and falls back to reasoning-based scheduling only when no suitable tool is available or the request falls outside tool coverage. 
In this fallback mode, physician schedules retrieved from FHIR are included in the prompt, allowing the agent to infer the earliest feasible slot consistent with the patient’s preferences.

In addition to new appointment scheduling, the staff agent also manages existing patient schedules. 
When a patient inquiry arrives, the staff agent infers the patient’s intent through dialogue and invokes the appropriate tool for rescheduling (moving appointments earlier) or cancellation.
In both cases, the agent first retrieves the original appointment; if retrieval fails due to missing information, the staff agent asks the necessary details (\eg full name).
For rescheduling, the appointment is reassigned if an eligible slot is available; otherwise, the patient is added to a waiting list.
For cancellation, once the appointment is canceled and a slot becomes free, the staff agent iterates through the waiting list and automatically reassigns appointments to patients who can be moved forward.

Simulation outcomes are evaluated using rubrics that assess schedule validity, alignment with patient preferences, and the success of rescheduling and cancellation.
Appendix~\ref{appendix:prompts:scheduling} provides prompts, and Appendix~\ref{appendix:simulation details:simulation_dialogues} presents example simulation dialogues.

\subsubsection{Hospital Information System Update via FHIR}
After all tasks are completed, new \textit{Patient} and \textit{Appointment} FHIR resources are created from the extracted patient information and final schedule.
Schedule changes---rescheduling or cancellation---are handled by updating or deleting the relevant \textit{Appointment} resources.
All FHIR operations are executed in real time throughout the simulation.

\section{Experiments}
\subsection{Experimental Setup}
\subsubsection{Hospital-Level Data Synthesis}
\label{sec:hospital_level_data_synthesis}
\input{tables/hospital_systhesis_level}

Experiments were conducted on synthesized hospital datasets at three levels---primary, secondary, and tertiary---each reflecting distinct institutional characteristics.
Higher-level hospitals include more departments and physicians and use finer time units, whereas primary hospitals involve patients with fewer prior diagnoses and weaker physician preferences during scheduling.
For each level, three datasets were generated for a 7-day simulation, yielding 516, 769, and 5,052 synthesized patient profiles for primary, secondary, and tertiary hospitals, respectively.

Table~\ref{tab:hospital_level_synthesis} summarizes the key synthesis parameters.
Among these, time granularity is an important design choice, as it directly affects scheduling resolution and physician capacity.
For tertiary hospitals, the time unit was set to 0.05 hours~\citep{Irving2017-rt}\footnote{Some countries report extremely short outpatient consultation times.} to reflect shorter outpatient consultations, whereas a coarser 0.25-hour unit was used for primary and secondary hospitals.
Physician capacity was derived from the time unit and randomly assigned within the allowable range for secondary and tertiary hospitals.
Hospital scale was controlled mainly through the number of departments and physicians: all nine internal medicine specialties were included for tertiary hospitals, while primary and secondary hospitals were assigned fewer departments (2--3 and 7--9, respectively).
Similarly, the number of physicians per department increased with the hospital level, ranging from one in primary hospitals to up to three in tertiary hospitals.
Other synthesis parameters were tuned to reflect typical differences across hospital levels.
In higher-level hospitals, patients were more likely to have prior diagnoses from lower-level institutions and to express explicit physician preferences.
Conversely, primary hospitals have longer outpatient working days for physicians and more $asap$ visits, with patients often arriving without prior appointments.
Scheduling preference types were assigned hierarchically: the primary preference was first sampled according to the distribution in Table~\ref{tab:hospital_level_synthesis}, and the secondary preference was then randomly selected from the remaining two types.

\subsubsection{Simulation Parameters}
During the intake simulation, the dialogue between the patient and administrative staff remains brief, as it does not involve physician-led clinical consultation.
Accordingly, each dialogue-based simulation (\eg intake and scheduling) is capped at five rounds, with each round consisting of a two-turn interaction between the patient and the staff.
For intake simulations, the current round index is injected into the staff agent’s system prompt, enabling it to generate contextually appropriate questions and finalize the department in the last round.
In the appointment scheduling simulation, rescheduling and cancellation event probabilities are set to 15\% and 10\% for tertiary hospitals---reflecting higher appointment demand and more tentative bookings---and 10\% and 5\% for primary and secondary hospitals, respectively.

\subsection{Evaluation Rubrics}
\label{sec:evaluation_rubrics}
The simulation quality and task outcomes are evaluated using carefully designed rubrics.
The rubrics consist of 10 evaluation criteria and 13 error codes, with some criteria associated with multiple error codes. 
Among them, 2 criteria and 2 error codes are shared between the intake and scheduling tasks. 
The intake task uses a total of 3 criteria and 5 error codes (Table~\ref{tab:error_analysis_intake}), 
and the scheduling task uses 9 criteria and 10 error codes (Table~\ref{tab:error_analysis_scheduling}).
Among the intake criteria, one evaluates the patient agent, while the others evaluate the staff agent.
The criteria are evaluated sequentially in a predefined order, and the evaluation terminates as soon as an error is detected, since several criteria depend on the outcomes of earlier ones.
A task is deemed successful only if no error is triggered and all criteria in the sequence are satisfied.
In particular, success in the intake task requires that no errors are triggered in either the patient or staff criteria.
Detailed rubrics---including evaluation flow, error codes, and suspected causes---are provided in Appendix~\ref{appendix:experimental details:rubrics}.

\subsection{Hospital-Level Simulation Results}
\label{sec:hospital-level simulation results}
\input{tables/main_results_extended}
\input{tables/error_analysis_intake_extended}
\input{tables/error_analysis_dept}
\mine enables a systematic analysis of LLM-based hospital administration across hospital levels.
We investigate how the hospital level affects task difficulty and the performance characteristics of LLMs.

Nine hospital datasets---three each for primary, secondary, and tertiary levels---were used for the administrative simulation.
Representative LLMs---GPT-5 Mini, GPT-5 Nano~\citep{gpt5}, Gemini 2.5 Flash~\citep{gemini2.5}, Llama 3.3 70B~\citep{grattafiori2024llama3herdmodels}, and Qwen 3 32B~\citep{yang2025qwen3technicalreport}---were evaluated, with the reasoning option set to $low$ for the GPT-5 series, $dynamic$ (default) for Gemini, and the default reasoning configuration for Qwen 3.
Both the patient and staff agents were run with the same model to measure model-specific task success rates.
Table~\ref{tab:simulation results} presents success rates for the patient intake and scheduling tasks, where the scheduling task includes rescheduling and cancellation events.
The scheduling task was evaluated under two settings: \textit{Scheduling (T)} (default), which prioritizes tool-calling for appointment scheduling and falls back to a reasoning-based mechanism upon failure, and \textit{Scheduling (R)}, which performs the scheduling process without any tools (\ie purely through reasoning).
Detailed analyses of the intake and scheduling results are provided in Sections~\ref{sec:hospital-level simulation results:intake analysis} and~\ref{sec:hospital-level simulation results:scheduling analysis}, respectively.
Lastly, the average token usage and API-related costs per simulation run are summarized in Appendix~\ref{appendix:token_cost}.

\subsubsection{Intake Analysis}
\label{sec:hospital-level simulation results:intake analysis}
\input{tables/error_analysis_scheduling}
Hospital-level differences in intake performance are largely driven by whether patients have prior diagnoses and by the number of departments, both of which strongly influence department assignment accuracy.
Other rubric criteria (\eg patient information extraction) are independent of hospital-level characteristics and instead depend on the model's intrinsic capabilities. 
In Table~\ref{tab:simulation results}, tertiary hospitals achieved higher intake success rates than secondary hospitals due to a larger proportion of patients with prior diagnoses. 
In contrast, primary hospitals achieved the highest success rates, even though they had the lowest proportion of patients referred with prior diagnoses. 
This is because the staff agent faced less ambiguity in department choices than in tertiary hospitals, which have up to four times more candidate departments.

Regarding model performance, Gemini 2.5 Flash achieved the highest success rates, whereas GPT-5 Mini underperformed relative to GPT-5 Nano. 
Among open-source models, Llama 3.3 70B achieved performance comparable to commercial models, ranking second in primary and tertiary hospital simulations, while Qwen 3 30B showed the lowest performance across all settings.
Table~\ref{tab:error_analysis_intake} shows the distribution of five intake error codes across three evaluation criteria.
In the staff evaluation, all models frequently exhibited department assignment errors (\textit{ID}+\textit{IDPI}), whereas formatting errors (\textit{IF}) and patient information extraction failures (\textit{IPI}+\textit{IDPI}) were rare.
In the patient evaluation, GPT-5 Mini showed a substantially higher rate of patient simulation failures (\textit{IS}) than other models, mainly due to the patient agent's failure to provide demographic information to the staff. 
Although GPT-5 Mini achieved the second-best staff performance, its poorer patient performance led to a lower overall intake success rate (Table~\ref{tab:simulation results}), as intake success requires both tasks to succeed.

Criterion \textit{C1-1} in Table~\ref{tab:error_analysis_intake} assesses output formatting.
If an error occurs at this stage, the intake task is classified as an \textit{IF} error, and the evaluation does not proceed to \textit{C1-2} (\ie department and information extraction), regardless of whether the department assignment is correct or incorrect (Section~\ref{sec:evaluation_rubrics}).
Therefore, we evaluate pure department assignment performance independently of the other rubric criteria, and report the results in Table~\ref{tab:error_analysis_dept}. 
We additionally record the number of dialogue rounds (out of five) required for each model to determine the department in dialogue with the patient, as a measure of the decision-making efficiency of the staff agent.
As this analysis ignores early termination in the rubric-based evaluation, department assignment errors previously masked by \textit{IF} become observable, increasing error counts compared to Table~\ref{tab:error_analysis_intake} in some cases.
Under this analysis, Gemini 2.5 Flash and GPT-5 Mini showed the fewest and second-fewest errors, respectively. 
Although GPT-5 Nano and Qwen 3 32B required the fewest rounds to reach decisions, their lower accuracy indicates an efficiency--accuracy trade-off.
Decisions tended to occur earlier in higher-level hospitals, and accuracy was lowest in secondary hospitals and highest in primary hospitals, consistent with the overall intake success trends shown in Table~\ref{tab:simulation results}.
Further human evaluation of dialogue quality and department-level misclassification patterns is provided in Appendices~\ref{appendix:human_evaluation} and~\ref{appendix:error patterns in department assignment}.

\subsubsection{Scheduling Analysis}
\label{sec:hospital-level simulation results:scheduling analysis}
\input{tables/tool_calling_success}
The scheduling task comprises three types of requests: new appointment scheduling, rescheduling of existing appointments, and appointment cancellation. 
Among these, rescheduling and cancellation require retrieval of the patient’s original appointment, making tool usage indispensable. 
However, new appointment scheduling and rescheduling after retrieval can be performed either through tool-based execution or pure LLM reasoning, corresponding to the \textit{Scheduling (T)} and \textit{Scheduling (R)} settings in Table~\ref{tab:simulation results}. 
Table~\ref{tab:error_analysis_scheduling} shows how the 10 scheduling error codes are distributed across nine evaluation criteria under the \textit{Scheduling (T)} setting.
The denominator of the \textit{Scheduling (T)} task exceeds the number of patient profiles described in Section~\ref{sec:hospital_level_data_synthesis}, because it incorporates both the one-to-one new appointment scheduling attempts for each patient profile and additional scheduling errors produced during rescheduling and cancellation events.

In Table~\ref{tab:simulation results} under the \textit{Scheduling (T)} setting, Gemini 2.5 Flash achieves the highest overall performance, followed by GPT-5 Mini, with both models exhibiting relatively stable performance across hospital levels.
GPT-5 Nano, Llama 3.3 70B, and Qwen 3 32B also exhibited comparable performance levels.
In Table~\ref{tab:error_analysis_scheduling}, \textit{FI} errors occur in rescheduling and cancellation requests when the original appointment cannot be retrieved, typically because the staff agent fails to elicit sufficient detail from the patient (\eg the physician’s full name).
All other error codes are directly related to scheduling outcomes, and the causes of their occurrence can be categorized into two types:
(1) cases where the agent misinterprets the patient's intent and invokes an inappropriate tool (\textit{Type I}); and
(2) cases where the agent judges that the request \textit{“cannot be handled by tools”} and falls back to reasoning-based scheduling, but the reasoning fails (\textit{Type II}), corresponding to \textit{Wrong Tool} and \textit{Fallback Failures} in Table~\ref{tab:tool_calling_analysis}.

In Table~\ref{tab:tool_calling_analysis}, rescheduling and cancellation events, which occur randomly during the simulation, are excluded to ensure a fair comparison.
Among the commercial models, GPT-5 Mini shows the highest rate of fallback transitions (\textit{Fallback Occurrences}), most frequently judging the available tools to be insufficient, suggesting a conservative tendency in interpreting patient preference utterances.
However, it exhibits the lowest failure rate among fallback cases, even though fallback failures are still observed.
This trend is consistent with Table~\ref{tab:simulation results}, where Mini achieves the lowest error rate under the \textit{Scheduling (R)} (reasoning-based) setting.
In other words, GPT-5 Mini's errors in Table~\ref{tab:error_analysis_scheduling} are not primarily driven by incorrect tool selection (\textit{Type I}), but reflect a small number of failures observed during frequent fallback reasoning (\textit{Type II}).
By contrast, in Table~\ref{tab:tool_calling_analysis}, GPT-5 Nano rarely judges that a request \textit{“cannot be handled by tools,”} resulting in fewer fallback transitions than Mini.  
Instead, most of GPT-5 Nano’s errors arise from incorrect tool invocation (\textit{Wrong Tool}), and even when fallback occurs, Nano shows a higher failure rate than GPT-5 Mini.
This suggests that the majority of GPT-5 Nano’s scheduling errors in Table~\ref{tab:error_analysis_scheduling} are driven by wrong tool selection rather than failures in fallback reasoning (\textit{Type I}).
This shows that GPT-5 Nano has weaker recognition of patient scheduling preferences in utterances, resulting in more frequent incorrect tool selection.
As shown in Table~\ref{tab:tool_calling_analysis}, Gemini 2.5 Flash neither invoked an incorrect tool nor triggered the fallback mechanism for new appointment requests, resulting in the most stable performance in the \textit{Scheduling (T)}.
For open-source models, most failures stem from incorrect tool selection (\textit{Type I}) rather than fallback failures (\textit{Type II}). 
When fallback is triggered, Qwen 3 32B exhibits a lower failure rate than Llama 3.3 70B, suggesting stronger reasoning-based scheduling performance, consistent with the results in Table~\ref{tab:simulation results} under the \textit{Scheduling (R)} setting.
Finally, detailed error analysis for the \textit{Scheduling (R)} setting is provided in Appendix~\ref{appendix:reasoning_error_analysis}.

\subsection{Factors Affecting Intake Performance}
\begin{figure}[t!]
    \includegraphics[width=\columnwidth]{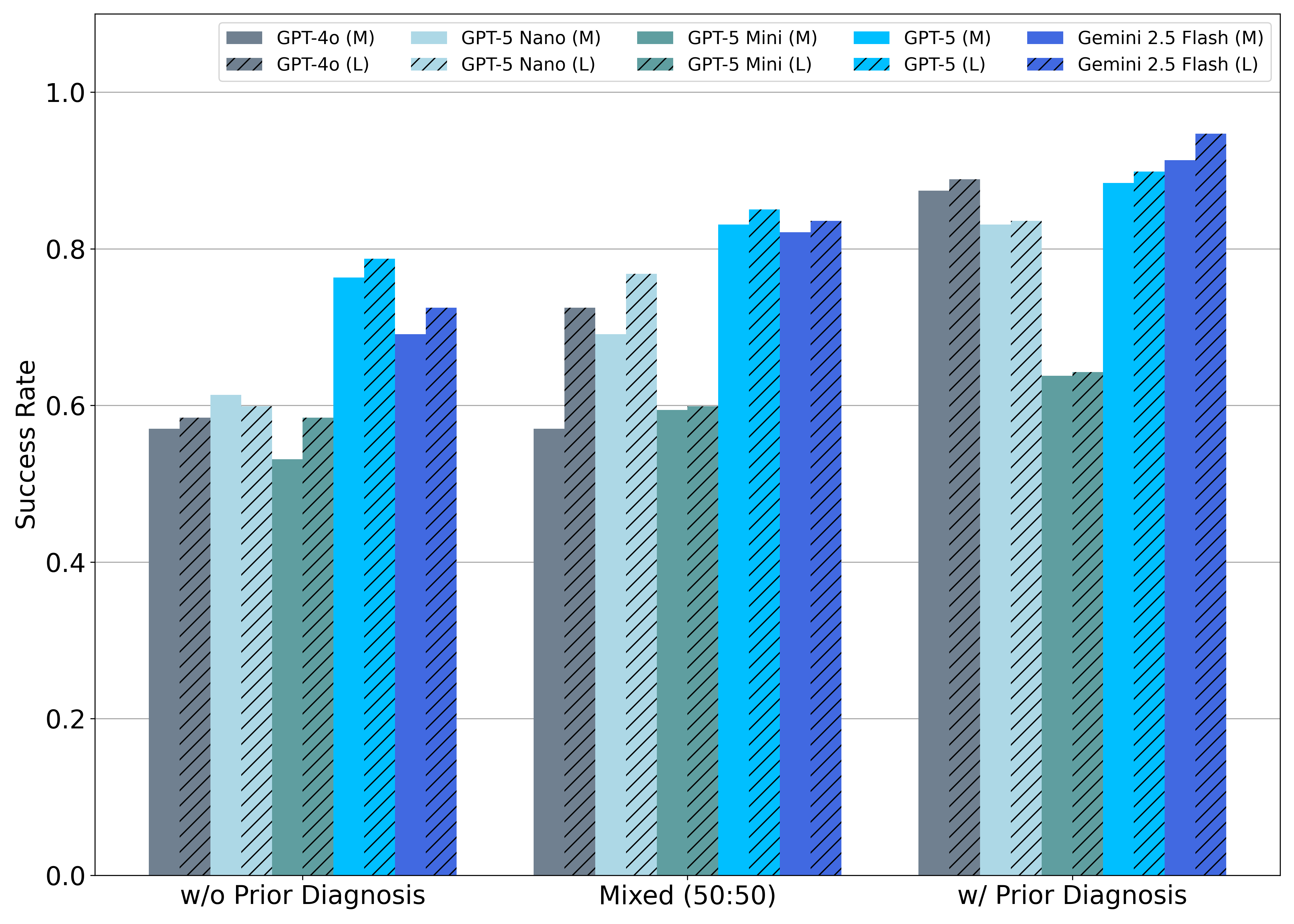}
    \caption{Comparison of intake task success rates across models under varying prior-diagnosis settings. Performance improves with higher prior-diagnosis rates and longer patient-staff dialogues (hatched bars).}
    \label{fig:intake_task_performance_anlysis1}
\end{figure}
\begin{figure}[t!]
    \includegraphics[width=\columnwidth]{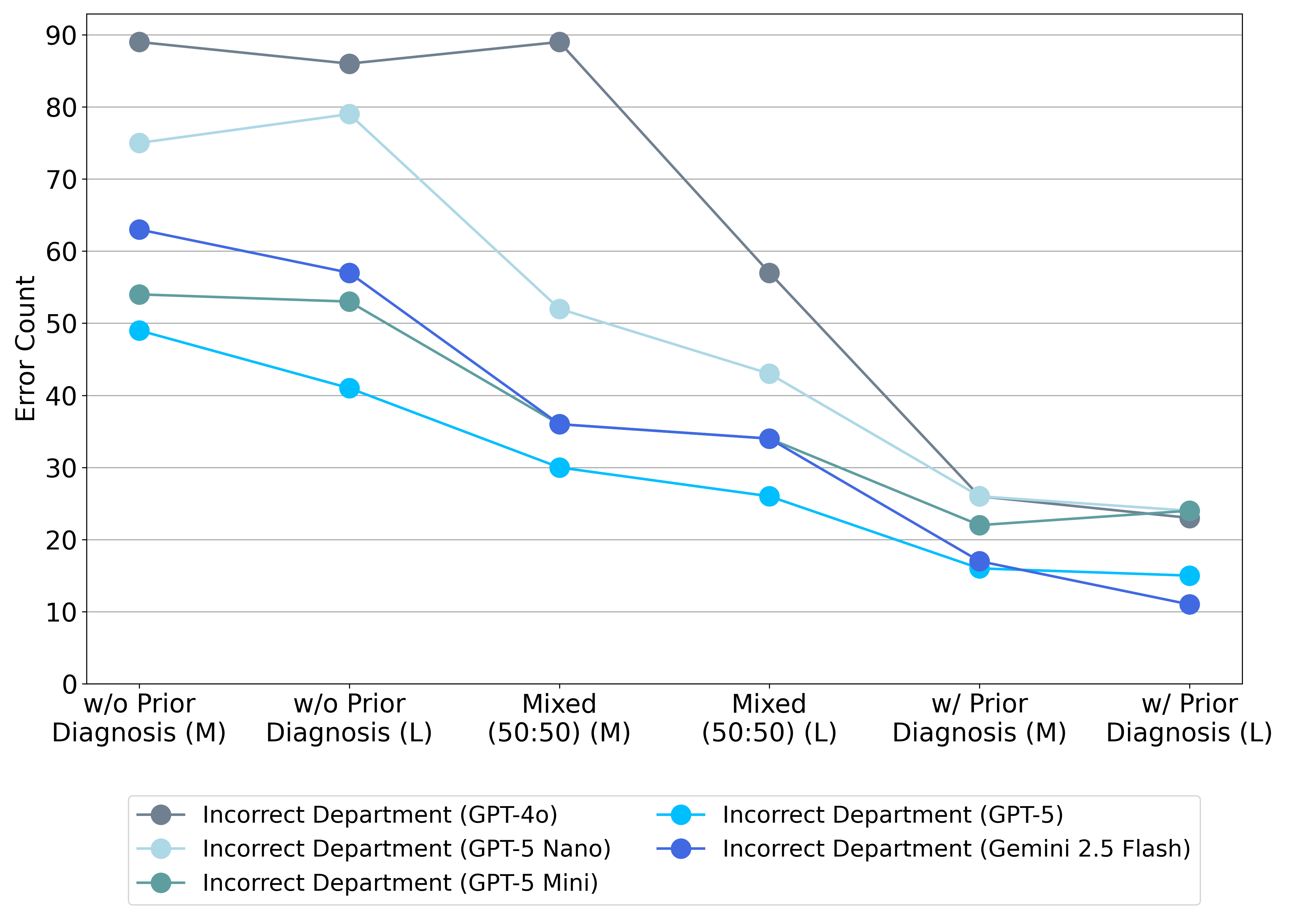}
    \caption{Department assignment errors under six conditions (three prior-diagnosis settings × two conversation-round settings). Errors decrease with more patients having prior diagnoses and with longer conversations.}
    \label{fig:intake_task_performance_anlysis2}
\end{figure}
While tool-based scheduling achieves high accuracy as long as appropriate tools are implemented (Table~\ref{tab:simulation results}), patient intake remains the primary bottleneck for LLM-based hospital administration.
As observed in Section~\ref{sec:hospital-level simulation results:intake analysis}, intake performance varies substantially across hospital levels depending on the availability of prior diagnoses and the model.
To examine these effects more explicitly and explore conditions under which intake performance can be improved, we constructed three datasets: (1) all patients without prior diagnosis; (2) 50\% with prior diagnosis; and (3) all with prior diagnosis, while keeping all other conditions identical.
Each dataset includes patients across all nine internal medicine departments.
Experiments were conducted using five models: GPT-4o~\citep{chatgpt4omini}, GPT-5 Nano, GPT-5 Mini, GPT-5, and Gemini 2.5 Flash.
We also investigated the effect of conversation length by comparing simulations with maximum limits of five rounds (default) and eight rounds.

Figure~\ref{fig:intake_task_performance_anlysis1} compares intake task success rates across models.
For all models, success rates increased when patients had prior diagnoses, which support more accurate department recommendations.
% Longer patient--staff dialogues also slightly improved performance, suggesting that three additional dialogue rounds before department decision help refine symptom descriptions.
Longer patient--staff dialogues also slightly improved performance, suggesting that three additional dialogue rounds before department decision help refine symptom descriptions and enable additional clarification through follow-up questions.
Figure~\ref{fig:intake_task_performance_anlysis2} further analyzes errors in department assignment only, under six conditions that combine three prior-diagnosis settings and two conversation-round settings.
The figure reports only department error counts, independent of the rubric evaluation order.
The results indicate that both prior diagnoses and longer dialogues consistently reduce department assignment errors across models.
In cases where the department had to be assigned based solely on symptoms (\ie without prior diagnoses), GPT-5 achieved the lowest department assignment error rates, whereas Gemini 2.5 Flash performed comparably when prior diagnoses were available.
These findings suggest that model choice, enriching prior-diagnosis information, and allowing slightly longer intake dialogues are all effective levers for improving LLM-based intake performance.

\section{Conclusion}
\label{sec:conclusion}
% This study presents \mine, a hospital administration simulation framework that supports synthetic data generation and end-to-end simulation of first-visit outpatient workflows with LLM agents. 
% We constructed detailed patient profiles using 194 labeled disease--symptom pairs across nine internal medicine specialties and diverse scheduling preferences, and modeled institution-specific characteristics when generating data and configuring simulation environments. 
% In addition, we provided a FHIR-integrated environment to support interoperability across institutions with heterogeneous data schemas.
% \mine serves as a unified testbed for hospital administrative workflows, enabling systematic evaluation of LLM-based administrative agents under diverse institutional settings. 
% Our results show that, when appropriate tools are available, tool-based scheduling delivers stable performance across hospital levels, whereas patient intake remains the primary bottleneck. 
% Intake performance improves markedly with stronger models and when patients have prior diagnoses, resulting in more accurate department assignment and higher overall intake success.
% These findings indicate that the two key requirements for reliable LLM-driven hospital administration are robust tool-based scheduling pipelines and patient-triage logic that determines which patients can be safely and effectively handled by LLM-based intake.
This study presents \mine, a hospital administration simulation framework for synthetic data generation and end-to-end simulation of first-visit outpatient workflows with LLM agents. 
We construct detailed patient profiles using 194 labeled disease--symptom pairs across nine internal medicine specialties and model institution-specific characteristics in FHIR-integrated environments. 
\mine serves as a unified testbed for evaluating LLM-based administrative agents under diverse institutional settings. 
Our results show that tool-based scheduling achieves stable performance when appropriate tools are available, whereas patient intake remains the primary bottleneck. 
These findings suggest that, for reliable LLM-driven administration, robust tool-based scheduling pipelines are essential, and careful patient allocation for LLM-based intake is required.

\section*{Limitation and Ethical Discussion}
\label{sec:limitation_and_ethical_discussion}
\noindent \textbf{Study Scope and Generalizability}. 
This study focuses on simulating administrative workflows specifically for first-time outpatient visits. 
Future research should extend this framework to encompass diverse follow-up scenarios and more intricate administrative processes to ensure broader clinical applicability.
\noindent \textbf{Safety and Data Privacy}. 
To ensure ethical rigor, all patient profiles and clinical data used in this study were synthetically generated, containing no real-world identifiable medical information. 
Thus, the dataset poses no risk of privacy infringement. 
However, practical deployment in clinical environments necessitates strict adherence to institutional data governance and privacy regulations. 
For API-based LLM services, robust safeguards---including data de-identification, stringent access control, and continuous auditing---must be implemented. 
Alternatively, deploying local LLMs within an on-premise hospital infrastructure can significantly mitigate external data leakage risks. 
Even when using models trained on real patient data, particularly when real-world patient data are collected via the simulator for training purposes, risk mitigation strategies such as differential privacy and output post-processing (\eg PII filtering) should be mandated to uphold the highest standards of patient confidentiality.

\acks{
This work was supported by the Institute of Information \& Communications Technology Planning \& Evaluation (IITP) grants (No.RS-2019-II190075, No.RS-2025-02304967), the Korea Health Industry Development Institute (KHIDI) grant (No.RS-2025-02223382), and National Research Foundation of Korea (NRF) grant (NRF-2020H1D3A2A03100945, RS-2026-25484088), funded by the Korea government (MSIT, MOHW).
}

\bibliography{reference}

\clearpage
\appendix

\section{Data Synthesis Details}
\label{appendix:data synthesis details}
\subsection{Data Synthesis Process}
\label{appendix:data synthesis details:conditions}
\input{algorithm/synthesize_algorithm}
\begin{figure*}[th!]
    \begin{center}
        \includegraphics[width=1\textwidth]{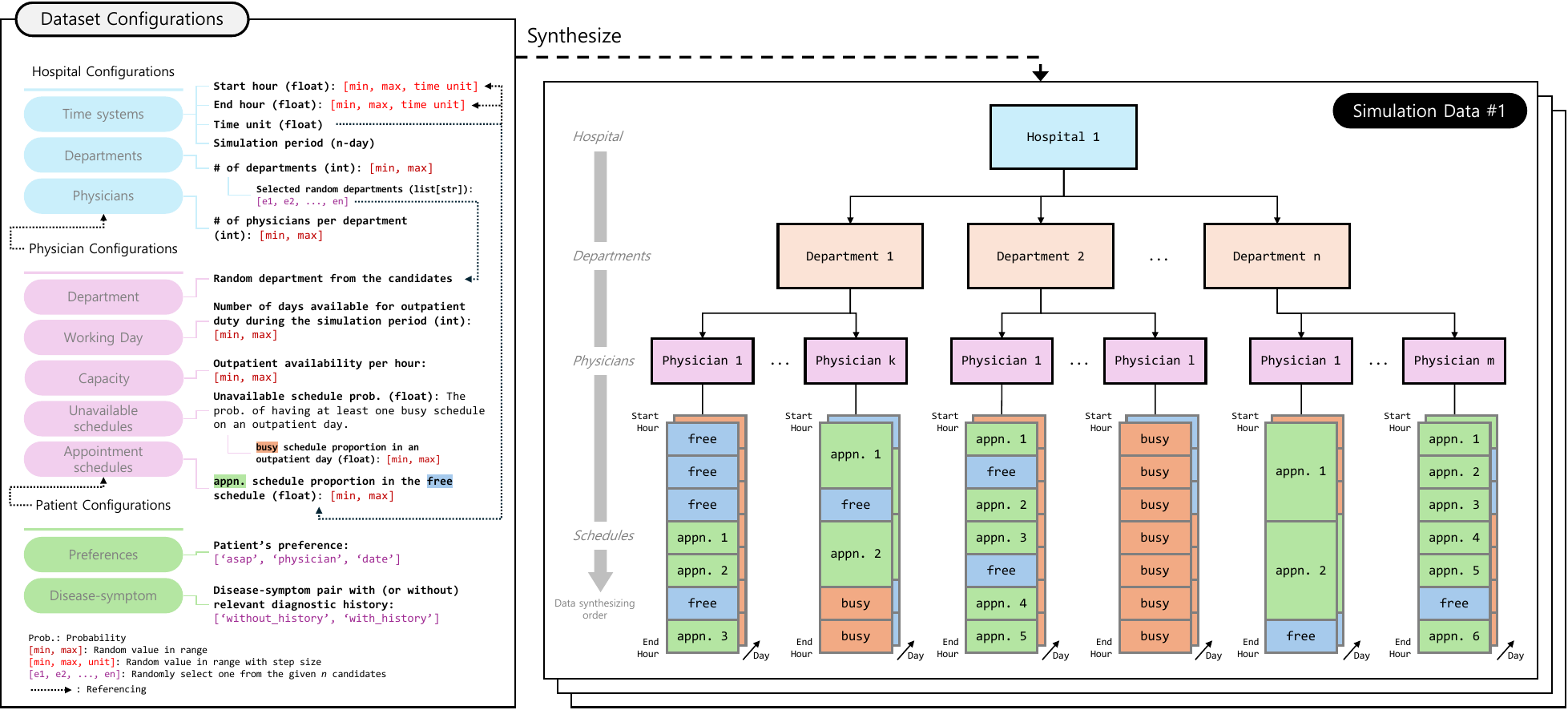}
        \caption{Overview of the hierarchical structure of the data synthesis procedure. The left panel presents the configuration parameters that govern data generation, and the right panel illustrates the hierarchical synthesis of hospital, department, physician, and patient data. Physician schedules consist of \textit{busy} and \textit{free} time slots, where \textit{free} slots represent periods available for outpatient appointments, and a subset of these \textit{free} slots is allocated as appointment blocks.}
        \label{fig:dataSynthesis}
    \end{center}
\end{figure*}
\input{appd_tables/synthesized_data}

The data synthesis procedure for the hospital administration simulation follows a hierarchical generation process with three stages (Algorithm~\ref{alg:synthetic-data}): hospital, physician, patient.
An overview of the hierarchical structure of the data synthesis procedure is illustrated in Figure~\ref{fig:dataSynthesis}.

\subsubsection{Hospital Data}
\label{appendix:hospital_data}
The synthesis of simulation data begins with generating hospital- and department-level information.
In this stage, users can specify parameters such as the number of hospitals to simulate, simulation period, start date, time unit ($\tau$), and operating hours.
A single $\tau$ corresponds to one discrete time interval (or time slot) within the hospital’s temporal system.
In addition, the number of departments and the number of physicians assigned to each department can be configured.
An example configuration is provided in the \textit{Hospital} section of Table~\ref{tab:synthesized data}.

\noindent\textbf{Departments}. 
We selected nine internal medicine specialties that are common across the specialty lists recognized by leading medical organizations in each country---the American College of Osteopathic Internists (United States),\footnote{\href{https://www.acoi.org/}{American College of Osteopathic Internists}} 
the General Medical Council (United Kingdom),\footnote{\href{https://www.gmc-uk.org/}{General Medical Council}} 
the Medical Board of Australia (Ahpra),\footnote{\href{https://www.medicalboard.gov.au/}{Medical Board of Australia}}
and the Canadian Medical Association (CMA)\footnote{\href{https://www.cma.ca/}{Canadian Medical Association}}
: gastroenterology, cardiology, pulmonology, endocrinology/metabolism, nephrology, hematology/oncology, allergy, infectious diseases, and rheumatology.

\subsubsection{Physician Data}
After hospital and department information are generated, physician data are synthesized according to the predefined number of physicians per department. 
Demographic attributes---including name, gender, birth date, and contact information (referred to as \textit{telecom} in FHIR)---are first created, followed by operational details such as working days, capacity, and schedules.
Working days are defined as days on which physicians are available for outpatient consultations; all time slots on working days are initialized as \textit{free}, whereas those on non-working days are marked as \textit{busy}.
The number of working days must not exceed the length of the simulation period. 
Physician capacity denotes the number of patients that can be consulted per hour, with the maximum capacity defined as $1/\tau$, where $\tau$ is the hospital time unit.
The actual capacity is randomly sampled from the divisors of this maximum, constrained by a predefined range.
Schedule synthesis incorporates probabilistic parameters. 
In Table~\ref{tab:synthesized data}, the parameter \textit{busy\_schedule\_prob} determines the likelihood that at least one \textit{busy} slot appears on a working day, and the proportion of \textit{busy} slots is sampled from the interval specified by \textit{busy\_schedule\_ratio}. 
A subset of the remaining \textit{free} slots is then assigned to appointments based on the \textit{appointment\_ratio}.
The size of an appointment block is determined by the physician’s capacity, resulting in varying slot durations across physicians.
Examples of physician-level synthesis parameters are provided in the \textit{Physician} section of Table~\ref{tab:synthesized data}.

\subsubsection{Patient Data}
\label{appendix:patient data}
After the total number of appointments is determined, patient profiles are generated and assigned to departments and physicians.
Demographic attributes---including name, gender, birth date, contact information (referred to as \textit{telecom} in FHIR), identifier, and address---are randomly synthesized. 
The identifier corresponds to a unique numeric value (\eg a national identification number) that distinguishes each patient.
Patient scheduling preferences are assigned probabilistically among three types: $asap$, $physician$, and $date$~\citep{douglas2005rheumatology,patrick2008dynamic,AGRAWAL20231318}. 
The $asap$ preference denotes scheduling the earliest available appointment with any physician within the designated department; the $physician$ preference specifies scheduling with a particular physician at their earliest availability; and the $date$ preference schedules the earliest appointment after a specified date.
These three preference types define primary and secondary scheduling preferences, where the \textit{preference} parameter in Table~\ref{tab:synthesized data} determines the primary type and the secondary type is randomly selected from the remaining two.

Each patient is also randomly assigned a disease--symptom profile consistent with their department. 
The presence of a prior diagnosis is assigned probabilistically according to the synthesis configuration.
Patients without a prior diagnosis are assumed to be unaware of their condition during the intake simulation. 
Relevant synthesis parameters are summarized in the \textit{Patient} section of Table~\ref{tab:synthesized data}, and the construction of disease--symptom pairs and their associated department labels is described in Appendix~\ref{appendix:disease labeling}.

\subsection{Construction of Disease--Symptom Datasets with Golden Department Labels}
\label{appendix:disease labeling}
\subsubsection{Obtaining Disease and Symptom Pairs}
\begin{figure*}[ht!]
    \begin{center}
        \scalebox{0.75}{\input{prompts/crawling_prompt}}
        \caption{System and user prompts used by the LLM for post-processing crawled data and extracting structured disease--symptom pairs. \texttt{\{DISEASE\}} and \texttt{\{WEBPAGE\}} denote placeholders for the disease name and its corresponding crawled webpage, respectively.}
        \label{fig:crawling_prompt}
    \end{center}
\end{figure*}
To simulate outpatient intake scenarios, disease--symptom profiles were constructed for patient profiles.
We used the Scotland National Health Service (NHS) website,\footnote{\href{https://www.nhsinform.scot/illnesses-and-conditions/a-to-z/}{NHS Inform Health Encyclopedia Scotland}.} which provides disease-centered information through A-to-Z health encyclopedias maintained by various national NHS entities.
Through web crawling, we collected 427 disease--symptom pairs, of which 194 corresponded to the nine internal medicine departments targeted in this study. 
The crawled data were subsequently post-processed using a Large Language Model (LLM) to extract symptom descriptions, yielding the final disease--symptom dataset.
GPT-4o~\citep{chatgpt4omini} was used for post-processing due to its efficiency, and the prompts employed for this step are illustrated in Figure~\ref{fig:crawling_prompt}.

\subsubsection{Golden Department Labeling}
\input{appd_tables/disease_labeling}
After constructing the disease--symptom dataset, golden labels were defined to indicate the medical departments responsible for treating each disease, serving as the reference standard for evaluating department assignment performance in the intake simulation.
All diseases were manually annotated using three authoritative publicly available clinical information sources: Seoul National University Hospital Medicine Information,\footnote{\href{https://www.snuh.org/health/nMedInfo/nList.do}{Seoul National University Hospital Medicine Information}} the Asan Medical Center Disease Encyclopedia,\footnote{\href{https://www.amc.seoul.kr/asan/healthinfo/disease/diseaseSubmain.do}{Asan Medical Center Disease Encyclopedia}} and the Severance Hospital Disease Encyclopedia.\footnote{\href{https://sev.severance.healthcare/health/encyclopedia/disease/disease.do}{Severance Hospital Disease Encyclopedia}}
Each disease was mapped to one or more of the nine internal medicine specialties.
Summary statistics and representative examples are provided in Table~\ref{tab:disease labeling results}. 
The total disease count reported in Table~\ref{tab:disease labeling results} exceeds 194 because diseases treatable across multiple departments are counted once for each applicable specialty.

\subsection{Examples of Synthetic Data}
\label{appendix:examples of synthetic data}
\begin{figure*}[ht!]
    \begin{center}
        \scalebox{0.75}{\input{appd_figures/synthetic_data_example}}
        \caption{Example of synthesized data used in the outpatient administration simulation, illustrating hospital (metadata), department information, and physician and patient profiles in JSON format.}
        \label{fig:synthesized data example}
    \end{center}
\end{figure*}
As illustrated in Figure~\ref{fig:synthesized data example}, the synthesized dataset for the simulation comprises metadata describing the simulation configuration, department information, physician profiles, and patient profiles with associated constraints and preferences.
The data are represented in JSON format, which preserves structured relationships and referential links among entities.

\section{FHIR Resources}
\label{appendix:fhir resource}
\begin{figure*}[ht!]
    \begin{center}
        \scalebox{0.75}{\input{appd_figures/fhir_resources}}
        \caption{Example FHIR resource instances used in the simulation.}
        \label{fig:fhir_examples}
    \end{center}
\end{figure*}
We adopt Fast Healthcare Interoperability Resources (FHIR) R5~\citep{6627810}, the latest version developed by Health Level Seven International.\footnote{\href{https://www.hl7.org/}{The official website of HL7 International}}
FHIR is a standardized framework for representing and exchanging healthcare information, enabling interoperability of electronic health records across institutions.
By leveraging web technologies (\ie RESTful APIs) and supporting JSON and XML formats, FHIR facilitates efficient and structured data exchange.
Accordingly, we integrate the FHIR standard into our simulation and summarize the key resources used in this work, with representative FHIR resource instances illustrated in Figure~\ref{fig:fhir_examples}.

\noindent\textbf{Practitioner}. 
The \textit{Practitioner} resource represents individuals who provide healthcare or related services and contains demographic attributes such as name, gender, birth date, and contact information.
In the simulation, this resource is used to represent each physician’s demographic profile.

\noindent\textbf{PractitionerRole}. 
The \textit{PractitionerRole} resource captures professional information about a practitioner, including workplace, specialty, and role-specific characteristics.
Here, it is used to record attributes such as department affiliation and outpatient capacity, which are essential for retrieving physician schedules and assigning appointments.

\noindent\textbf{Patient}. 
The \textit{Patient} resource stores demographic and administrative information about individuals receiving care, including name, gender, birth date, identifiers, and contact details.
This resource represents each simulated patient within the system.

\noindent\textbf{Schedule}. 
The \textit{Schedule} resource defines a practitioner’s general availability across multiple time periods.
Each schedule links to a \textit{Practitioner} and specifies the overall availability window without exposing individual appointment details, ensuring privacy.
In our simulation, one schedule is created per practitioner for the entire simulation period.

\noindent\textbf{Slot}. 
A \textit{Slot} subdivides the \textit{Schedule} into discrete time intervals defined by the simulation time unit ($\tau$).
Each slot references its parent \textit{Schedule} resource and indicates whether the practitioner is available or busy during that interval.
For example, with $\tau = 0.25$ hours, each slot corresponds to a 15-minute unit.
In this study, slots are marked as either \textit{free} or \textit{busy}, indicating whether the interval is available for appointment booking or already occupied.
This structure is essential for constructing the practitioner’s complete schedule throughout the simulation.

\noindent\textbf{Appointment}. 
The \textit{Appointment} resource represents an outpatient encounter between a patient and a physician.
Appointments reference one or more consecutive \textit{free} \textit{Slot} resources (depending on physician capacity) and link to both the \textit{Practitioner} and \textit{Patient} resources.
During the simulation, appointment creation, cancellation, and modification update physician schedules in real time.

\section{Experimental Details}
\label{appendix:experimental details}
\subsection{Evaluation Rubrics}
\label{appendix:experimental details:rubrics}
\begin{figure*}[ht!]
    \begin{center}
        \includegraphics[width=1\textwidth]{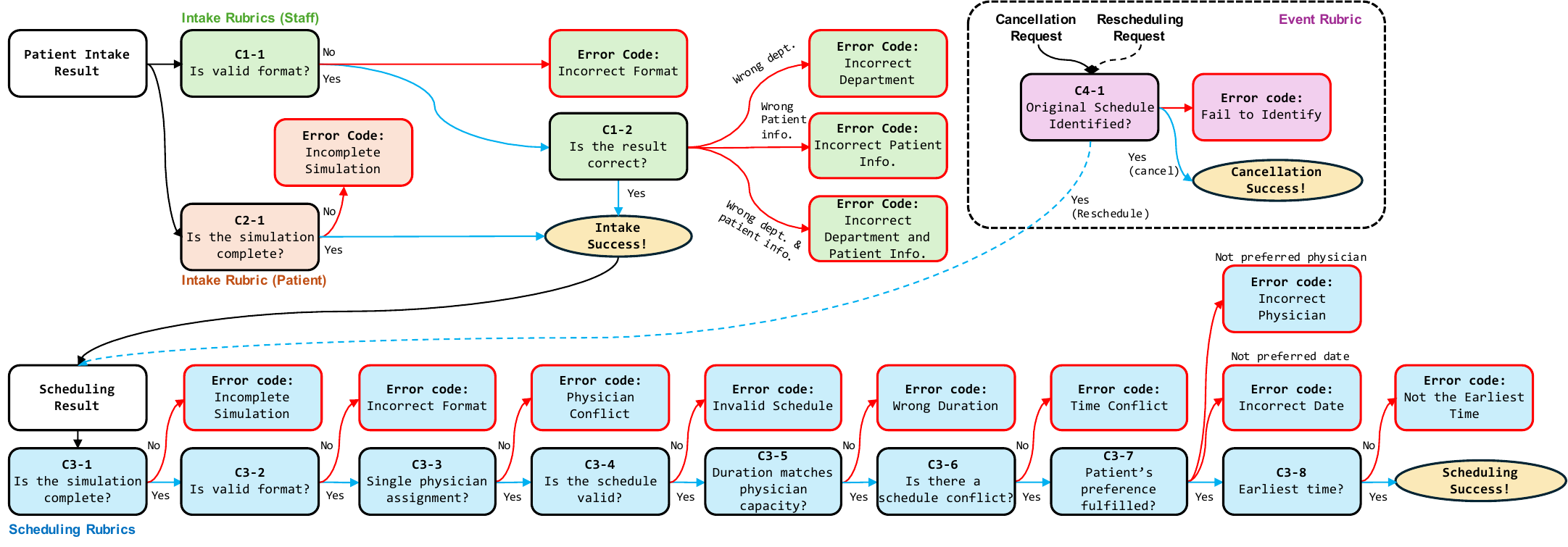}
        \caption{Procedure diagram of the evaluation rubrics. \textit{Abbreviation: Info. = Information; Dept. = Department}.}
        \label{fig:rubric_evaluation}
    \end{center}
\end{figure*}
\input{appd_tables/rubric_table}
The proposed framework evaluates task outcomes using predefined rubrics.
Each task rubric includes a set of criteria and associated error codes: the intake task uses three criteria with three task-specific error codes and two common error codes, while the scheduling task uses eight criteria with seven task-specific error codes and two common error codes.
Rescheduling and cancellation events each include one task-specific error code.
Among the intake criteria, one evaluates the patient agent, while the others evaluate the staff agent.
The criteria are evaluated sequentially in a predefined order, and the evaluation terminates as soon as an error is detected, since several criteria depend on the outcomes of earlier ones.
A task is deemed successful only if no error is triggered and all criteria in the sequence are satisfied.
Figure~\ref{fig:rubric_evaluation} shows the overall evaluation flow.

Regarding the intake task, the framework assesses both the patient and the staff agents. 
The staff evaluation focuses on adherence to the output format and the accuracy of information extraction and department assignment, while the patient evaluation ensures that the simulation is complete---\ie whether the agent faithfully conveys the pre-assigned role-play information. 
These two evaluations are performed independently, and the intake task is considered successful only if both evaluations are passed.
For the scheduling task, the evaluation examines whether the staff agent completes the task execution, and then assesses response formatting, the validity of the generated schedule, the appropriateness of the assigned physician, and whether patient preferences are properly reflected.
The framework also verifies whether the agent identifies the earliest feasible appointment slot while satisfying all scheduling constraints.
For the rescheduling and cancellation events, the rubric first checks whether the agent successfully retrieves the patient's requested original schedule and then follows different evaluation flows depending on the event type.
The full mapping between error codes and their corresponding suspected causes is provided in Table~\ref{tab:rubric table}.

\subsection{Token Usage and Operational Costs}
\label{appendix:token_cost}
\input{tables/token_cost}
This section analyzes token usage and the associated economic cost in simulations using API-based LLMs. 
Table~\ref{tab:token_cost} reports the average token consumption per simulation run and the corresponding API cost, aggregated over 50 runs for each task and model. 
Since rescheduling and cancellation events occur randomly and vary with the length of the waiting list, we restrict our analysis to the new appointment scheduling scenario to ensure consistency in resource usage measurements.

When considering deployment in real-world healthcare administrative settings, the primary focus is on the token consumption and cost of the staff agent, which serves as the core operator of the system. 
The results show that, for both GPT-5 Mini and Nano, the staff agent incurred the highest cost in the \textit{Intake} task, while no significant cost difference was observed between \textit{Scheduling (T)} and \textit{Scheduling (R)}. 
In particular, the cost of the intake task was up to 4.44$\times$ higher than scheduling for GPT-5 Mini and up to 2.05$\times$ higher for GPT-5 Nano.
In contrast, Gemini 2.5 Flash exhibited the highest cost in the \textit{Scheduling (R)} task. 
This is attributed to significantly increased reasoning token consumption when scheduling is performed without tool usage.
Finally, assuming each model performs an end-to-end task (\textit{Intake} + \textit{Scheduling (T)}), the average API cost per simulation run was highest for GPT-5 Mini (\$0.007652) and lowest for GPT-5 Nano (\$0.001953).

\subsection{Human Evaluation of Intake Dialogues}
\label{appendix:human_evaluation}
\begin{figure*}[th!]
    \begin{center}
        \includegraphics[width=1\textwidth]{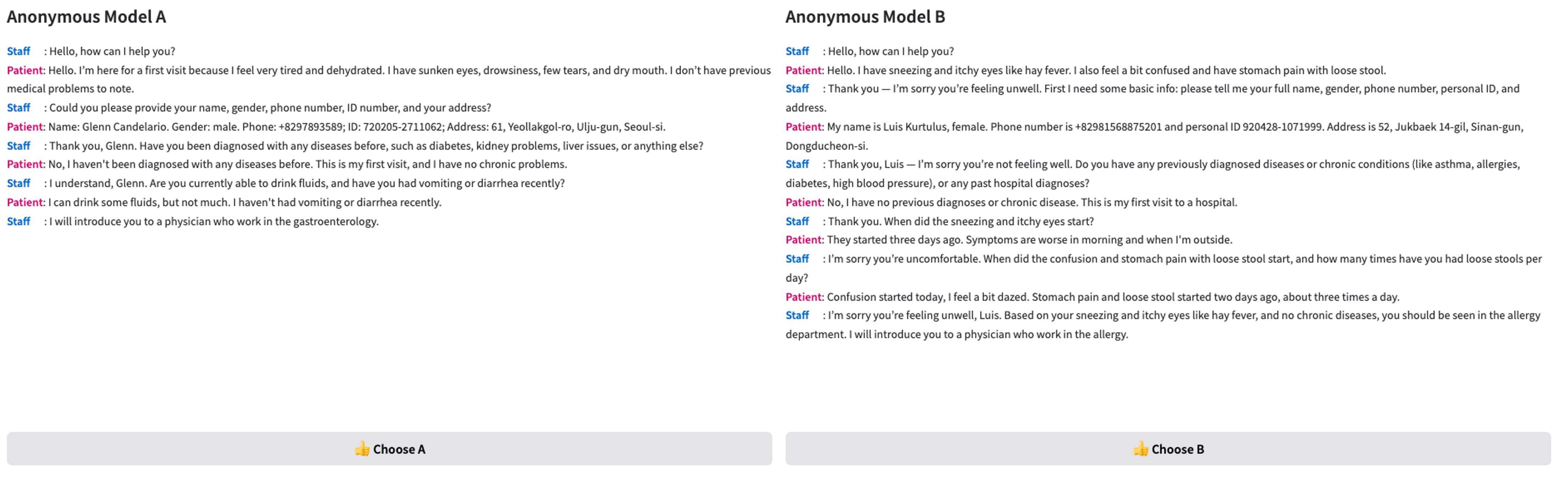}
        \caption{Human evaluation interface used for the intake-task dialogue assessment. Evaluators were presented with two anonymized dialogues generated by different models (GPT-5 Mini, GPT-5 Nano, Gemini 2.5 Flash) and asked to select the more realistic interaction (arena evaluation). After making a selection, they proceeded to score each dialogue using predefined criteria.}
        \label{fig:arena}
    \end{center}
\end{figure*}
\input{appd_tables/human_eval}

We conducted a human evaluation to qualitatively assess the patient-staff dialogues generated during the intake task.
As shown in Figure~\ref{fig:arena}, evaluators were given two anonymized dialogues produced by randomly selected models among GPT-5 Mini, GPT-5 Nano, and Gemini 2.5 Flash, and performed anonymous arena voting and scoring.
For the arena evaluation, evaluators selected the dialogue that more closely resembled a realistic interaction between a patient and hospital administrative staff.
Each dialogue was then rated on a 5-point scale according to four criteria, with one point deducted for each criterion that was not adequately satisfied:
(1) Patient: naturalness of symptom descriptions and avoidance of excessive medical jargon;
(2) Staff: whether the staff agent avoided clinician-like behavior (\eg diagnosing), asked appropriate intake-related questions, and maintained an empathetic and polite tone;
(3) Flow: whether the dialogue naturally covered essential intake steps (greeting, asking demographic information, symptom collection, prior-diagnosis verification, and department assignment) and achieved these objectives;
(4) Overall: clarity of language, naturalness of closing statements, and resemblance to a real hospital reception scenario.

A total of 17 evaluators participated, and results are summarized in Table~\ref{tab:human eval}.
Gemini 2.5 Flash achieved the strongest performance, with 89 arena wins and an average score of 4.11, followed by GPT-5 Nano with 80 wins and a score of 4.06.
GPT-5 Mini received the lowest ratings, primarily because it often adopted clinician-like behavior near the end of the conversation---offering suspected diagnoses or detailed clinical assessments---which frequently resulted in point deductions.

\subsection{Error Patterns in Department Assignment}
\label{appendix:error patterns in department assignment}
\begin{figure*}[th!]
    \begin{center}
        \includegraphics[width=1\textwidth]{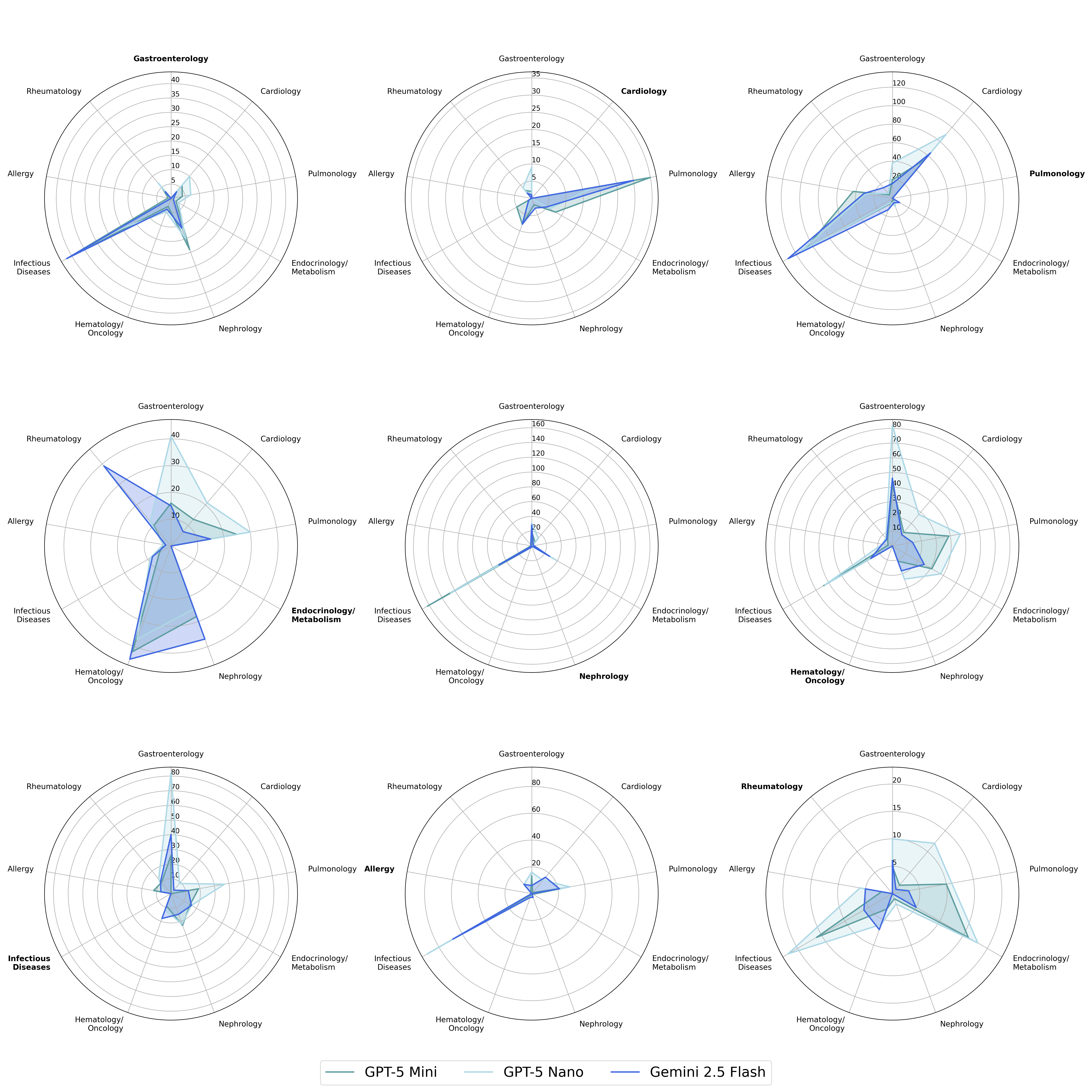}
        \caption{Figure~\ref{fig:dept_error_radar_all}. Department-level misclassification patterns across the nine internal medicine specialties. Each radar plot illustrates the frequency of incorrect department recommendations made by the staff agent for diseases belonging to that specialty. Bold labels indicate the ground-truth department for each plot.}
        \label{fig:dept_error_radar_all}
    \end{center}
\end{figure*}
This section analyzes misclassification patterns in the departments recommended by the staff agent, as illustrated in Figure~\ref{fig:dept_error_radar_all}.
Across most specialties, misclassification into infectious diseases occurred at a notably high rate.
This pattern arises because many non-specific symptoms---such as cough, sneezing, fever, runny nose, and sore throat---are common across multiple departments and are often associated with infections.
Meanwhile, certain infectious disease cases include gastrointestinal symptoms (\eg nausea, diarrhea, vomiting), leading to frequent misclassification as gastroenterology.
We also observe that cardiology and pulmonology are frequently confused with one another, likely because diseases in these specialties share overlapping symptoms such as chest pain and various respiratory complaints.
For endocrinology/metabolism, all three models showed frequent misclassification into hematology/oncology or nephrology. 
This pattern can be attributed to overlapping symptoms such as chronic fatigue and weight loss, which are characteristic of both metabolic disorders and various cancers or kidney diseases.
Finally, rheumatology was often misclassified not only as infectious diseases but also as endocrinology/metabolism.
Rheumatologic conditions present with a broad symptom spectrum---extending beyond muscle pain and joint symptoms to include hormone---related manifestations (\eg dry mouth, dry eyes) and systemic features such as chills and sleep disturbances---resulting in overlap with other departments.

\subsection{Reasoning-based Scheduling Error Analysis}
\label{appendix:reasoning_error_analysis}
\input{appd_tables/error_analysis_schedule_R}
In Table~\ref{tab:simulation results}, under the \textit{Scheduling (R)} setting, GPT-5 Mini achieved the highest success rate among the evaluated models. 
Despite tertiary hospitals having the smallest time units (\ie the most finely segmented time slots) and the largest numbers of departments and physicians---thus constituting the most challenging scenario---GPT-5 Mini still achieved a success rate of 75\%, whereas GPT-5 Nano, Gemini 2.5 Flash, Llama 3.3 70B, and Qwen 3 32B reached only 17.4\%, 37.9\%, 13.2\%, and 23.8\%, respectively.
These results indicate that GPT-5 Mini exhibits substantially stronger reasoning capability for scheduling than the other models.
Table~\ref{tab:error_analysis_scheduling_R} summarizes the error distribution for the \textit{Scheduling (R)} setting. 
Except for GPT-5 Mini, other models produced a wide variety of errors; the most frequent was failure to choose the earliest feasible time slot (\textit{NET}), largely due to difficulty in identifying mid-schedule vacant slots created by cancellations.

In contrast to the general trend observed across all other models, where the lowest success rates occurred in tertiary hospital settings, Gemini 2.5 Flash performed worst in the primary hospital setting, even though we originally expected tertiary hospitals to be the most challenging.
This pattern arises because Gemini 2.5 Flash has difficulty identifying the earliest available slot for $asap$ preferences, regardless of physician or date constraints. 
In particular, it often misinterprets chronological order---for example, treating a slot at 9:00 on September 11, 2025, as earlier than one at 13:00 on September 10, 2025---which makes scheduling for patients with an $asap$ preference more challenging than for those specifying a $physician$ or $date$. 
Because primary hospitals operate at a smaller scale and face lower outpatient competition, most patients prefer the earliest available appointment, and the primary-hospital data therefore contain roughly 20\% more patients with $asap$ preferences than the secondary- or tertiary-hospital data (Table~\ref{tab:hospital_level_synthesis}). 
This higher prevalence of $asap$ preferences makes the issue particularly evident in the primary-hospital results (Table~\ref{tab:simulation results}): Gemini 2.5 Flash exhibits a 63.3\% error rate in primary-level scheduling simulations, slightly higher than in tertiary-level simulations (62.5\%).
Consistent with this pattern, the most frequent errors are \textit{NET} (\textit{Not the Earliest Time}), arising from failures to assign the earliest available slots.

Finally, criteria \textit{C3-7} and \textit{C3-8} evaluate how accurately the agent captures the patient’s scheduling preferences expressed in their utterances. 
More specifically, \textit{Incorrect Physician (IP)} assesses whether the appointment is made with the physician preferred by the patient, \textit{Incorrect Date (IDT)} assesses whether the patient’s date constraints (\eg “after a specific date”) are satisfied, and \textit{Not the Earliest Time (NET)} evaluates whether the scheduled appointment is the earliest available slot that satisfies the preceding constraints or, in cases with no additional constraints, whether the request for the earliest possible appointment is correctly fulfilled.
As shown in Table~\ref{tab:error_analysis_scheduling_R}, \textit{IP} and \textit{IDT} errors are the least frequent across all models and scenarios, whereas \textit{NET} errors are by far the most frequent.
This suggests that, while the agents are generally successful at inferring basic intent and minimal requirements from patient utterances (\eg preferred physician and date constraints), they still exhibit weaknesses in identifying the earliest feasible time slots on the schedule.

\section{Simulation Details}
\label{appendix:prompts}
This section provides the full prompt templates used for the patient intake and appointment scheduling tasks.

\subsection{Patient Intake}
\label{appendix:prompts:intake}
\subsubsection{Patient Persona Details}
\label{appendix:persona}
\input{appd_tables/persona_desc}
To realistically simulate outpatient intake in the administrative department, patient personas must be carefully defined.
In this study, we used the disease profiles described in Section~\ref{sec:patient conditions} and Appendix~\ref{appendix:disease labeling}, along with additional persona attributes that capture patient personality traits.
\textsc{PatientSim}~\citep{kyung2025patientsimpersonadrivensimulatorrealistic} supports four configurable personality-related dimensions:
(1) personality,
(2) language proficiency,
(3) confusion level, and
(4) medical history recall level.
Patient agents were configured to reflect typical outpatient characteristics.
Among the six available personality types (neutral, distrustful, impatient, overanxious, overly positive, and verbose), we used the \textit{neutral} type.
Language proficiency was set to the \textit{intermediate} level (level B on the Common European Framework of Reference for Languages, CEFR A–C scale)~\citep{council2001common}, representing moderate medical comprehension.
The confusion level was set to \textit{normal}, the lower of the two available levels (normal and high), as most outpatient visits are non-emergent.
For patients with prior diagnoses, the medical history recall level was set to \textit{high}; otherwise, it was configured as \textit{no\_history}.
Because \textsc{PatientSim} was originally designed for emergency-department scenarios and did not include a \textit{no\_history} option, we extended the framework to support outpatient use.
Descriptions of each personality type are summarized in Table~\ref{tab:persona desc} and applied to the patient-agent prompts shown in Figure~\ref{fig:intake_patient_prompt}.

\subsubsection{Patient Prompt}
The patient agent assumes it is interacting with the hospital’s administrative staff, as specified in the intake prompt. 
As shown in Figure~\ref{fig:intake_patient_prompt}, the prompt consists of three main components: (1) patient information, (2) persona, and (3) behavioral guidelines. 
The patient information section includes demographic details (\eg name, gender, phone number) and prior diagnosis. 
If no prior diagnosis exists, the \texttt{\{diagnosis\}} and \texttt{\{medical\_history\}} fields are set to “unknown” and “first hospital visit for this symptom,” respectively. 
Conversely, if the patient has a prior diagnosis, these fields are populated with the corresponding disease name and relevant diagnostic information. 
The \texttt{\{chief\_complaint\}} field lists the symptoms currently experienced by the patient. 
The persona component is constructed by choosing settings from \textsc{PatientSim} that best match outpatient scenarios (Appendix~\ref{appendix:persona}). 
Additionally, the persona descriptions summarized in Table~\ref{tab:persona desc} are incorporated into the \texttt{\{reminder\}} field to reinforce consistent patient behavior during the interaction. 
The behavioral guideline component integrates the instructions shown in Figure~\ref{fig:intake_patient_guideline}, ensuring coherent conversational behavior throughout the simulation. 
Since lengthy utterances are uncommon in outpatient settings, the \texttt{\{sentence\_limit\}} value is set to 3. 
For extensibility, optional fields such as \texttt{\{allergies\}} and \texttt{\{family\_medical\_history\}} are also supported.

\subsubsection{Administrative Staff Prompt}
Figure~\ref{fig:intake_staff_prompt} illustrates the prompt used for the administrative staff role in the intake simulation.
The staff agent’s task is to gather patient information through dialogue and assign each patient to an appropriate medical department based on their reported symptoms and any available prior diagnosis.
The guideline first instructs the agent to verify demographic information and check for prior diagnostic records, reflecting standard outpatient procedures.
It is designed to minimize technical terminology, enabling the agent to elicit the patient’s symptoms and chief complaint in a natural and patient-friendly manner.
Because administrative consultations are generally brief, \texttt{\{total\_idx\}} is set to five rounds (a 10-turn dialogue), with \texttt{\{curr\_idx\}} and \texttt{\{remain\_idx\}} updated in real time to support contextually appropriate questioning.
Finally, the hospital’s department list \texttt{\{department\}} is referenced when generating the final department recommendation.

Figure~\ref{fig:intake_staff_task_prompt} presents the user prompt used for post-dialogue processing. 
After completing the conversation while retaining full contextual information, the staff agent is instructed to extract the patient’s demographic details and the finalized department recommendation. 
This prompt converts the dialogue-derived information into a structured format for use in subsequent tasks, ensuring consistency and facilitating downstream evaluation.

\subsection{Appointment Scheduling}
\label{appendix:prompts:scheduling}
\input{appd_tables/preference_desc}
\subsubsection{Patient Prompt}
Once the patient agent is assigned to a clinical department after the intake process, it engages in a conversation with the administrative staff agent to schedule an appointment. 
Figure~\ref{fig:schedule_patient_prompt} illustrates the system prompt used by the patient agent to express its primary scheduling preference during the interaction with the staff agent. 
In this prompt, \texttt{\{preference\}} denotes the primary preference type selected from the three predefined preference categories, while \texttt{\{preference\_desc\}} corresponds to the description associated with the selected preference type as defined in Table~\ref{tab:preference_desc}. 
In addition, \texttt{\{preferred\_doctor\}} contains the physician assigned during the data generation phase when the preference type is \textit{physician}, and is set to “N/A” for all other preference types.
With a probability of 30\%, the patient agent rejects the initial schedule proposed by the staff agent and requests an alternative based on its secondary preference. 
In such cases, the original system prompt is replaced with the prompt shown in Figure~\ref{fig:schedule_reject_patient_prompt}, where the preference-related components are updated to reflect the secondary preference type. 
Furthermore, \texttt{\{rejected\_preference\}} contains the description of the rejected primary preference.
Finally, \texttt{personality} is set to the \textit{neutral} type, consistent with the intake stage.

In addition, \mine supports not only new patient appointment scheduling but also simulations of rescheduling and cancellation requests for existing appointments.
Figure~\ref{fig:reschedule_patient_prompt} and Figure~\ref{fig:cancel_patient_prompt} present the system prompts of the patient agent for rescheduling and cancellation, respectively.
Each prompt includes the patient's personality along with information about the currently scheduled appointment, enabling the agent to generate rescheduling or cancellation requests based on this information.

\subsubsection{Administrative Staff Prompt}
The administrative agent primarily handles patient requests through tool invocation. 
Figure~\ref{fig:tool_staff_prompt} presents the tool-calling prompt used by the staff agent. 
This prompt is designed to analyze the patient's utterance and select an appropriate tool among scheduling, rescheduling, and cancellation tasks simulated in this study. 
In particular, the scheduling-related tools include descriptions corresponding to the three scheduling preference types supported in our framework. 
In addition, when required information for tool invocation is missing or a tool call fails, the staff agent can ask clarification questions to obtain more accurate information from the patient. 
Furthermore, if errors occur during scheduling or if it cannot be performed using the available tools, a “NO TOOL” flag is returned to trigger LLM-based reasoning for scheduling as a fallback mechanism. 
This safety mechanism is designed to enable flexible responses even in real-world scenarios where predefined tools are not available.

Figure~\ref{fig:schedule_staff_prompt} presents the prompt used for the fallback mechanism in LLM-based reasoning for scheduling.
The prompt consists of three components: (1) hospital information, (2) patient information, and (3) physician information.
First, the hospital information includes operating hours, time unit, and virtual current time in the simulation. 
Second, the patient information---comprising the assigned department (\texttt{\{DEPARTMENT\}}) and utterance of scheduling preference (\texttt{\{PREFERENCE\}})---provides the basis for determining appropriate scheduling actions. 
When LLM-based reasoning is triggered for rescheduling, the \texttt{\{RESCHEDULING\_FLAG\}} is used to instruct the model to thoroughly search for available time slots.
Finally, physician information---provided via the \texttt{\{DOCTOR\}} placeholder---includes schedule tables for the entire simulation period. 
The consolidated scheduling rules derived from this process are shown in Figure~\ref{fig:schedule_staff_guideline} and inserted into the \texttt{\{SCHEDULING\_RULES\}} placeholder.

\subsection{Example Simulation Dialogues}
\label{appendix:simulation details:simulation_dialogues}
Figure~\ref{fig:simulation_dialogues} presents example dialogues generated in our simulation.
In the \textit{Intake + Scheduling} examples, the first scenario illustrates a patient with prior diagnoses who changes their scheduling preference during the interaction. The second scenario shows how the staff agent assigns the patient to an appropriate department based solely on symptom information in the absence of prior diagnoses, and then schedules an appointment according to the patient’s primary scheduling preference.
In the rescheduling and cancellation examples, the staff agent infers the patient’s intent from the utterance and invokes the appropriate tools to complete rescheduling or cancellation requests. In particular, the rescheduling scenario also demonstrates how the staff agent asks follow-up questions to obtain missing information required for tool invocation.

\begin{figure*}[ht!]
    \centering
    \scalebox{0.75}{\input{prompts/intake_patient_system_prompt}}
    \caption{System prompt template for role-playing the patient in the intake simulation. Braced elements (\{\}) are substituted with patient-specific values.}
    \label{fig:intake_patient_prompt}
\end{figure*}
\begin{figure*}[ht!]
    \centering
    \scalebox{0.75}{\input{prompts/intake_patient_guideline}}
    \caption{The prompt represents the general behavioral guideline for a patient during the intake simulation. This guideline is inserted into the \texttt{\{behavioral\_guideline\}} placeholder in Figure~\ref{fig:intake_patient_prompt}.}
    \label{fig:intake_patient_guideline}
\end{figure*}

\begin{figure*}[ht!]
    \centering
    \scalebox{0.75}{\input{prompts/intake_staff_system_prompt}}
    \caption{System prompt template for role-playing the administrative staff in the intake simulation. Braced elements (\{\}) are replaced with the list of hospital departments and the current dialogue-round information; the values of \texttt{\{curr\_idx\}} and \texttt{\{remain\_idx\}} are dynamically updated to generate contextually appropriate questions for each round.}
    \label{fig:intake_staff_prompt}
\end{figure*}
\begin{figure*}[ht!]
    \centering
    \scalebox{0.75}{\input{prompts/intake_staff_task_user_prompt}}
    \caption{Prompt used by the administrative staff agent for post-dialogue processing in the intake task. The prompt instructs the agent to extract key patient information and return it in a structured JSON format.}
    \label{fig:intake_staff_task_prompt}
\end{figure*}
\begin{figure*}[ht!]
    \centering
    \scalebox{0.75}{\input{prompts/schedule_patient_system}}
    \caption{Patient agent system prompt template for new appointment scheduling.}
    \label{fig:schedule_patient_prompt}
\end{figure*}
\begin{figure*}[ht!]
    \centering
    \scalebox{0.75}{\input{prompts/schedule_reject_patient_system}}
    \caption{Patient agent system prompt template for rejecting the initial schedule proposal.}
    \label{fig:schedule_reject_patient_prompt}
\end{figure*}
\begin{figure*}[ht!]
    \centering
    \scalebox{0.75}{\input{prompts/reschedule_patient_system}}
    \caption{Patient agent system prompt template for appointment rescheduling.}
    \label{fig:reschedule_patient_prompt}
\end{figure*}
\begin{figure*}[ht!]
    \centering
    \scalebox{0.75}{\input{prompts/cancel_patient_system}}
    \caption{Patient agent system prompt template for appointment cancellation.}
    \label{fig:cancel_patient_prompt}
\end{figure*}
\begin{figure*}[ht!]
    \centering
    \scalebox{0.75}{\input{prompts/tool_staff_prompt}}
    \caption{Tool-calling prompt for the administrative staff agent.}
    \label{fig:tool_staff_prompt}
\end{figure*}
\begin{figure*}[ht!]
    \centering
    \scalebox{0.75}{\input{prompts/schedule_staff_prompt}}
    \caption{System and user prompts for the administrative staff agent in the appointment scheduling task.}
    \label{fig:schedule_staff_prompt}
\end{figure*}
\begin{figure*}[ht!]
    \centering
    \scalebox{0.75}{\input{prompts/schedule_staff_guideline}}
    \caption{The prompt represents the scheduling rules for the administrative staff agent during the appointment scheduling simulation. These rules are inserted into the \texttt{\{SCHEDULING\_RULES\}} placeholder in Figure~\ref{fig:schedule_staff_prompt}.}
    \label{fig:schedule_staff_guideline}
\end{figure*}
\begin{figure*}[ht!]
    \centering
    \scalebox{0.75}{\input{appd_figures/dialogues}}
    \caption{Example dialogues from the hospital administration simulation.}
    \label{fig:simulation_dialogues}
\end{figure*}

\end{document}

%% file: tables/hospital_systhesis_level.tex
\begin{table*}[!th]
    \caption{Summary of the synthesized hospital dataset configuration for the 7-day simulation. For each hospital level, three hospital instances were randomly generated, and values in square brackets ([]) indicate sampling ranges. \textit{Abbreviations: Dept. = department, h = hours, N = number, P = probability}.}
    \label{tab:hospital_level_synthesis}
    \begin{center}
        \resizebox{1.0\linewidth}{!}{
            \begin{tabular}{lrrrrrrr}
            \toprule
            \textbf{Hospital Level} & \textbf{Time Unit (h)} & \textbf{\textit{N} of Dept.} & \textbf{Physicians per Dept.} & \textbf{Working Days} & \textbf{Capacity} & \makecell[r]{\textbf{\textit{P} of First Priority Preference} \\ (asap, physician, date)} & \makecell[r]{\textbf{\textit{P} of Prior Diagnosis} \\ (w/o, w/)} \\
            \midrule
            \textbf{Primary}    & 0.25 & [2, 3] & [1, 1] & [5, 7] & [4, 4] & (0.6, 0.2, 0.2) & (0.9, 0.1) \\
            \textbf{Secondary}  & 0.25 & [7, 9] & [1, 2] & [3, 4] & [1, 4] & (0.4, 0.4, 0.2) & (0.6, 0.4) \\
            \textbf{Tertiary}   & 0.05 & [9, 9] & [2, 3] & [3, 4] & [1, 20] & (0.4, 0.4, 0.2) & (0.2, 0.8) \\
            \bottomrule
            \end{tabular}
        }
    \end{center}
\end{table*}

%% file: tables/main_results_extended.tex
\begin{table}[t!]
    \caption{Success rate comparison across hospitals and models. \textit{Scheduling (T)} uses tool-calling, while \textit{Scheduling (R)} uses LLM reasoning only. Shaded cells indicate open-source models. Bold and underlined values denote the best and second-best performance, respectively.}
    \label{tab:simulation results}
    \resizebox{1.0\linewidth}{!}{
        \begin{tabular}{llrr|r}
        \toprule
        \multirow[c]{2}{*}{\textbf{Hospital}} & \multirow[c]{2}{*}{\textbf{Model}} & \multicolumn{3}{c}{\textbf{Task (\%)}} \\
        \cmidrule(lr{2pt}{2pt}){3-5}
        & & \makecell[c]{\textbf{Intake}} & \makecell[c]{\textbf{Scheduling (T)}} & \makecell[c]{\textbf{Scheduling (R)}} \\
        \midrule
        \multirow[t]{3}{*}{\textbf{Primary}}   & GPT-5 Mini       & 0.670$\pm$3.67 & \textbf{0.998$\pm$0.31} & \textbf{0.990$\pm$0.27} \\
                                               & GPT-5 Nano       & 0.791$\pm$6.61 & 0.967$\pm$0.52 & 0.556$\pm$10.04 \\
                                               & Gemini 2.5 Flash & \textbf{0.889$\pm$2.77} & \textbf{0.998$\pm$0.22} & 0.371$\pm$5.59 \\
        \rowcolor{gray!10}
                             \cellcolor{white} & Llama 3.3 70B    & \underline{0.818$\pm$5.34} & 0.941$\pm$0.98 & 0.402$\pm$7.35 \\
        \rowcolor{gray!10}
                             \cellcolor{white} & Qwen 3 32B       & 0.615$\pm$0.28 & 0.939$\pm$4.08 & \underline{0.594$\pm$3.16} \\
        \midrule
        \multirow[t]{3}{*}{\textbf{Secondary}} & GPT-5 Mini       & 0.570$\pm$1.34 & \underline{0.998$\pm$0.17} & \textbf{0.967$\pm$1.73} \\
                                               & GPT-5 Nano       & \underline{0.723$\pm$0.48} & 0.951$\pm$0.79 & 0.672$\pm$8.31 \\
                                               & Gemini 2.5 Flash & \textbf{0.812$\pm$0.82} & \textbf{0.999$\pm$0.12} & 0.660$\pm$5.18\\
        \rowcolor{gray!10}
                             \cellcolor{white} & Llama 3.3 70B    & 0.706$\pm$4.74 & 0.932$\pm$2.33 & 0.341$\pm$11.37 \\
        \rowcolor{gray!10}
                             \cellcolor{white} & Qwen 3 32B       & 0.528$\pm$4.14 & 0.963$\pm$1.34 & \underline{0.699$\pm$2.18} \\
        \midrule
        \multirow[t]{3}{*}{\textbf{Tertiary}}  & GPT-5 Mini       & 0.613$\pm$0.79 & \underline{0.970$\pm$0.59} & \textbf{0.750$\pm$1.45} \\
                                               & GPT-5 Nano       & 0.754$\pm$0.95 & 0.933$\pm$2.38 & 0.174$\pm$0.75 \\
                                               & Gemini 2.5 Flash & \textbf{0.853$\pm$0.89} & \textbf{0.996$\pm$0.28} & \underline{0.379$\pm$3.36} \\
        \rowcolor{gray!10}
                             \cellcolor{white} & Llama 3.3 70B    & \underline{0.768$\pm$0.77} & 0.814$\pm$4.74 & 0.132$\pm$1.34 \\
        \rowcolor{gray!10}
                             \cellcolor{white} & Qwen 3 32B       & 0.576$\pm$0.46 & 0.929$\pm$1.07 & 0.238$\pm$3.67 \\
        \bottomrule
        \end{tabular}
        }
\end{table}

%% file: tables/error_analysis_intake_extended.tex
\begin{table*}[t!]
    \caption{Error distribution in the intake task by hospital level, model, and error code. The prefixes of \textit{Intake (Staff)} (\eg \textit{C1-1}, \textit{C1-2}) denote the staff agent evaluation criteria and the order in which they are applied. Criterion \textit{C1-2} can produce one of three error codes and \textit{C2-1} of \textit{Intake (Patient)} evaluates the completeness of the patient agent's simulation (Appendix~\ref{appendix:experimental details:rubrics}). Shaded cells indicate open-source models. Bold and underlined values denote the best and second-best performance, respectively. \textit{Abbreviations: IF = Incorrect Format, ID = Incorrect Department, IPI = Incorrect Patient Information, IDPI = Incorrect Department \& Patient Information, IS = Incomplete Simulation.}}
    \label{tab:error_analysis_intake}
    \small
    \begin{center}
    \resizebox{1\linewidth}{!}{
        \begin{tabular}{llr||rrrrr}
        \toprule
        \multirow[c]{2}{*}{\textbf{Hospital}} & \multirow[c]{2}{*}{\textbf{Model}} & \multicolumn{5}{c}{\textbf{Intake (Staff)}} & \multicolumn{1}{c}{\textbf{Intake (Patient)}} \\

        \cmidrule{3-7}
        \cmidrule(lr{2pt}{2pt}){3-7} \cmidrule(lr{2pt}{2pt}){8-8}
        & & \makecell[c]{\textbf{Errors}} & \makecell[c]{\textbf{C1-1. IF}} & \makecell[c]{\textbf{C1-2. ID}} & \makecell[c]{\textbf{C1-2. IPI}} & \makecell[c]{\textbf{C1-2. IDPI}} & \makecell[c]{\textbf{C2-1. IS}} \\
        \midrule
        \multirow[t]{5}{*}{\textbf{Primary}}   & GPT-5 Mini       & \underline{61 / 516 (11.8\%)} & 1 (1.6\%) &  59 (96.7\%) & 1 (1.6\%) & - & 127 / 516 (24.6\%) \\
                                               & GPT-5 Nano       & 88 / 516 (17.1\%) & 2 (2.3\%) &  85 (96.6\%) & 1 (1.1\%) & - & 33 / 516 (6.4\%) \\
                                               & Gemini 2.5 Flash & \textbf{58 / 516 (11.2\%)} & - & 58 (100.0\%) & - & - & \textbf{2 / 516 (0.4\%)} \\
        \rowcolor{gray!10}
                             \cellcolor{white} & Llama 3.3 70B    & 96 / 516 (18.6\%) & - & 86 (89.6\%) & 6 (6.3\%) & 4 (4.2\%) & \underline{14 / 516 (2.7\%)} \\
        \rowcolor{gray!10}
                             \cellcolor{white} & Qwen 3 32B       & 106 / 516 (20.5\%) & 1 (0.9\%) & 100 (94.3\%) & 5 (4.7\%) & - & 120 / 516 (23.3\%) \\
        \midrule
        \multirow[t]{5}{*}{\textbf{Secondary}} & GPT-5 Mini       & \underline{152 / 769 (19.8\%)} & - & 150 (98.7\%)  & 2 (1.3\%) & - & 233 / 769 (30.3\%)\\
                                               & GPT-5 Nano       & 193 / 769 (25.1\%) & 2 (1.0\%) & 189 (97.9\%) & 1 (0.5\%) & 1 (0.5\%) & 27 / 769 (3.5\%)\\
                                               & Gemini 2.5 Flash & \textbf{143 / 769 (18.6\%)} & - & 142 (99.3\%) & - & 1 (0.7\%) & \textbf{2 / 769 (0.3\%)} \\
        \rowcolor{gray!10}
                             \cellcolor{white} & Llama 3.3 70B    & 228 / 769 (29.7\%) & - & 218 (95.6\%) & 6 (2.6\%) & 4 (1.8\%) & \underline{10 / 769 (1.3\%)} \\
        \rowcolor{gray!10}
                             \cellcolor{white} & Qwen 3 32B       & 203 / 769 (26.4\%) & 1 (0.5\%) & 198 (97.5\%) & 3 (1.5\%) & 1 (0.5\%) & 207 / 769 (26.9\%) \\
        \midrule
        \multirow[t]{5}{*}{\textbf{Tertiary}}  & GPT-5 Mini       & \underline{854 / 5,052 (16.9\%)} & 2 (0.2\%) & 850 (99.5\%) & 2 (0.2\%) & - & 1,347 / 5,052 (26.7\%) \\
                                               & GPT-5 Nano       & 1,068 / 5,052 (21.1\%) & 10 (0.9\%) & 1,046 (97.9\%) & 9 (0.8\%) & 3 (0.3\%) & 231 / 5,052 (4.6\%) \\
                                               & Gemini 2.5 Flash & \textbf{731 / 5,052 (14.5\%)} & - & 728 (99.6\%) & 2 (0.3\%) & 1 (0.1\%) & \textbf{7 / 5,052 (0.1\%)} \\
        \rowcolor{gray!10}
                             \cellcolor{white} & Llama 3.3 70B    & 1,170 / 5,052 (23.2\%) & - & 1,148 (98.1\%) & 14 (1.2\%) & 8 (0.7\%) & \underline{23 / 5,052 (0.5\%)} \\
        \rowcolor{gray!10}
                             \cellcolor{white} & Qwen 3 32B       & 992 / 5,052 (19.6\%) & 3 (0.3\%) & 961 (96.9\%) & 22 (2.2\%) & 6 (0.6\%) & 1,436 / 5,052 (28.4\%) \\
        \bottomrule
        \end{tabular}
    }
    \end{center}
\end{table*}

%% file: tables/error_analysis_dept.tex
\begin{table}[t!]
    \caption{Department assignment error rates and average dialogue rounds (maximum five rounds) in the intake task across hospital levels and models. Shaded cells indicate open-source models. Bold and underlined values denote the best and second-best performance, respectively.}
    \label{tab:error_analysis_dept}
    \small
    \begin{center}
    \resizebox{1\linewidth}{!}{
        \begin{tabular}{llrr}
        \toprule
        \textbf{Hospital} & \textbf{Model} & \textbf{Avg. Rounds} & \textbf{Department Errors} \\
        \midrule
        \multirow[t]{5}{*}{\textbf{Primary}}   & GPT-5 Mini       & 4.86$\pm$0.44 & \underline{59 / 516 (11.4\%)} \\
                                               & GPT-5 Nano       & \underline{3.78$\pm$0.65} & 85 / 516 (16.5\%) \\
                                               & Gemini 2.5 Flash & 4.34$\pm$0.74 & \textbf{58 / 516 (11.2\%)} \\
        \rowcolor{gray!10}
                             \cellcolor{white} & Llama 3.3 70B    & 3.86$\pm$0.36 & 90 / 516 (17.4\%) \\
        \rowcolor{gray!10}
                             \cellcolor{white} & Qwen 3 32B       & \textbf{2.78$\pm$0.67} & 101 / 516 (19.6\%) \\
        \midrule
        \multirow[t]{5}{*}{\textbf{Secondary}} & GPT-5 Mini       & 4.75$\pm$0.60 & \underline{150 / 769 (19.5\%)} \\
                                               & GPT-5 Nano       & \underline{3.70$\pm$0.66} & 191 / 769 (24.8\%) \\
                                               & Gemini 2.5 Flash & 4.07$\pm$0.88 & \textbf{143 / 769 (18.6\%)} \\
        \rowcolor{gray!10}
                             \cellcolor{white} & Llama 3.3 70B    & 3.71$\pm$0.46 & 222 / 769 (28.9\%) \\
        \rowcolor{gray!10}
                             \cellcolor{white} & Qwen 3 32B       & \textbf{2.67$\pm$0.70} & 200 / 769 (26.0\%) \\
        \midrule
        \multirow[t]{5}{*}{\textbf{Tertiary}}  & GPT-5 Mini       & 4.67$\pm$0.66 & \underline{850 / 5,052 (16.8\%)} \\
                                               & GPT-5 Nano       & \underline{3.54$\pm$0.70} & 1,055 / 5,052 (20.9\%) \\
                                               & Gemini 2.5 Flash & 3.62$\pm$0.82 & \textbf{729 / 5,052 (14.4\%)} \\
        \rowcolor{gray!10}
                             \cellcolor{white} & Llama 3.3 70B    & 3.55$\pm$0.50 & 1,156 / 5,052 (22.9\%) \\
        \rowcolor{gray!10}
                             \cellcolor{white} & Qwen 3 32B       & \textbf{2.53$\pm$0.72} & 970 / 5,052 (19.2\%) \\
        \bottomrule
        \end{tabular}
    }
    \end{center}
\end{table}

%% file: tables/error_analysis_scheduling.tex
\begin{table*}[t!]
    \caption{Error distribution in the tool-calling-based scheduling task by hospital level, model, and error code. The prefixes (\eg \textit{C3-1}, \textit{C3-2}) denote the evaluation criteria and the order in which they are evaluated, where criterion \textit{C3-7} can yield one of two error codes, and \textit{C4-1} applies only to rescheduling and cancellation (Appendix~\ref{appendix:experimental details:rubrics}). Shaded cells indicate open-source models. Bold and underlined values denote the best and second-best performance, respectively. \textit{Abbreviations: IS = Incomplete Simulation, IF = Incorrect Format, PC = Physician Conflict, IVS = Invalid Schedule, WD = Wrong Duration, TC = Time Conflict, IP = Incorrect Physician, IDT = Incorrect Date, NET = Not the Earliest Time, FI = Fail to Identify.}}
    \label{tab:error_analysis_scheduling}
    \begin{center}
        \resizebox{1\linewidth}{!}{
            \begin{tabular}{llr||rrrrrrrrr|r}
            \toprule
            \multirow[c]{2}{*}{\textbf{Hospital}} & \multirow[c]{2}{*}{\textbf{Model}} & \multicolumn{11}{c}{\textbf{Scheduling (T)}} \\
            \cmidrule{3-13}
            & & \makecell[c]{\textbf{Errors}} & \makecell[c]{\textbf{C3-1. IS}} & \makecell[c]{\textbf{C3-2. IF}} & \makecell[c]{\textbf{C3-3. PC}} & \makecell[c]{\textbf{C3-4. IVS}} & \makecell[c]{\textbf{C3-5. WD}} & \makecell[c]{\textbf{C3-6. TC}} & \makecell[c]{\textbf{C3-7. IP}} & \makecell[c]{\textbf{C3-7. IDT}} & \makecell[c]{\textbf{C3-8. NET}} & \makecell[c]{\textbf{C4-1. FI}} \\
            \midrule
            \multirow[t]{5}{*}{\textbf{Primary}}   & GPT-5 Mini       & \textbf{1 / 596 (0.2\%)}  & - & - & - & - & - & - & - & - & 1 (100.0\%) & - \\
                                                   & GPT-5 Nano       & 20 / 590 (3.4\%) & - & - & - & - & - & - & - & 19 (95.0\%) & 1 (5.0\%) & - \\
                                                   & Gemini 2.5 Flash & \textbf{1 / 582 (0.2\%)}  & - & - & - & - & - & - & - & - & - & 1 (100.0\%) \\
            \rowcolor{gray!10}
                                 \cellcolor{white} & Llama 3.3 70B & 34 / 571 (6.0\%)  & - & 1 (2.9\%) & - & - & - & 1 (2.9\%) & - & 4 (11.8\%) & 4 (11.8\%) & 24 (70.6\%) \\
            \rowcolor{gray!10}
                                 \cellcolor{white} & Qwen 3 32B & 30 / 577 (5.2\%)  & 1 (3.3\%) & - & - & - & - & 1 (3.3\%) & - & 14 (46.7\%) & 7 (23.3\%) & 7 (23.3\%) \\
            \midrule
            \multirow[t]{5}{*}{\textbf{Secondary}} & GPT-5 Mini       & \underline{2 / 898 (0.2\%)}  & - & - & - & - & - & - & - & 2 (100.0\%) & - & - \\
                                                   & GPT-5 Nano       & 46 / 908 (5.1\%) & - & - & - & - & 1 (2.2\%) & - & 1 (2.2\%) & 42 (91.3\%) & 2 (4.4\%) & - \\
                                                   & Gemini 2.5 Flash & \textbf{1 / 905 (0.1\%)}  & - & - & - & - & - & - & - & - & - & 1 (100.0\%) \\
            \rowcolor{gray!10}
                                 \cellcolor{white} & Llama 3.3 70B & 55 / 859 (6.4\%)  & - & - & - & - & - & - & - & 13 (23.6\%) & 4 (7.3\%) & 38 (69.1\%) \\
            \rowcolor{gray!10}
                                 \cellcolor{white} & Qwen 3 32B & 28 / 858 (3.3\%)  & 1 (3.6\%) & - & - & - & - & 1 (3.6\%) & 3 (10.7\%) & 3 (10.7\%) & 9 (32.1\%) & 11 (39.3\%) \\
            \midrule
            \multirow[t]{5}{*}{\textbf{Tertiary}}  & GPT-5 Mini       & \underline{202 / 6,736 (3.0\%)} & 1 (0.5\%) & 27 (13.4\%) & - & - & 14 (6.9\%) & 2 (1.0\%) & 6 (3.0\%) & 12 (5.9\%) & 140 (69.3\%) & - \\
                                                   & GPT-5 Nano       & 455 / 6,811 (6.7\%) & 2 (0.4\%) & 81 (17.8\%) & - & - & 15 (3.3\%) & 19 (4.2\%) & 16 (3.5\%) & 229 (50.3\%) & 92 (20.2\%) & 1 (0.2\%) \\
                                                   & Gemini 2.5 Flash & \textbf{25 / 6,550 (0.4\%)} & - & - & - & - & - & - & - & - & 23 (92.0\%) & 2 (8.0\%) \\
            \rowcolor{gray!10}
                                 \cellcolor{white} & Llama 3.3 70B & 1,174 / 6,441 (18.2\%)  & - & 1 (0.1\%) & - & 4 (0.3\%) & 1 (0.1\%) & 28 (2.4\%) & - & 109 (9.3\%) & 167 (14.2\%) & 864 (73.6\%) \\
            \rowcolor{gray!10}                                
                                 \cellcolor{white} & Qwen 3 32B & 454 / 6,747 (6.7\%)  & 2 (0.4\%) & 3 (0.7\%) & - & - & - & 62 (13.7\%) & 27 (6.0\%) & 81 (17.8\%) & 172 (37.9\%) & 107 (23.6\%) \\
            \bottomrule
            \end{tabular}
        }
    \end{center}
\end{table*}

%% file: tables/tool_calling_success.tex
\begin{table}[t!]
    \caption{Failure analysis in the \textit{Scheduling (T)} new appointment task.  \textit{Wrong Tool} indicates failures due to incorrect tool invocation, and \textit{Fallback Failures} refers to failures occurring during the reasoning-based scheduling process initiated by a fallback. Shaded cells indicate open-source models.}
    \label{tab:tool_calling_analysis}
    \small
    \begin{center}
    \resizebox{1\linewidth}{!}{
        \begin{tabular}{llrr|r}
        \toprule
        \multirow[c]{2}{*}{\textbf{Hospital}} & \multirow[c]{2}{*}{\textbf{Model}} & \multirow[c]{2}{*}{\textbf{Wrong Tool}} & \multicolumn{2}{c}{\textbf{Fallback}} \\
        \cmidrule(lr{2pt}{2pt}){4-5}
        & & & \makecell[c]{\textbf{Occurrences}} & \makecell[c]{\textbf{Failures}}\\
        \midrule
        \multirow[t]{5}{*}{\textbf{Primary}}   & GPT-5 Mini       & \textbf{0 / 516 (0.0\%)} & 4 / 516 (0.8\%) & \textbf{1 / 4 (25.0\%)} \\
                                               & GPT-5 Nano       & 19 / 516 (3.7\%) & 3 / 516 (0.6\%) & 1 / 3 (33.3\%) \\
                                               & Gemini 2.5 Flash & \textbf{0 / 516 (0.0\%)} & \textbf{0 / 516 (0.0\%)} & - \\
        \rowcolor{gray!10}
                             \cellcolor{white} & Llama 3.3 70B    & 7 / 516 (1.4\%) & 4 / 516 (0.8\%) & 3 / 4 (75.0\%) \\
        \rowcolor{gray!10}
                             \cellcolor{white} & Qwen 3 32B       & 30 / 516 (5.8\%) & 7 / 516 (1.4\%) & 2 / 7 (28.6\%) \\
        \midrule
        \multirow[t]{5}{*}{\textbf{Secondary}} & GPT-5 Mini       & \textbf{0 / 769 (0.0\%)} & 6 / 769 (0.8\%) & \textbf{2 / 6 (33.3\%)} \\
                                               & GPT-5 Nano       & 43 / 769 (5.6\%) & 3 / 769 (0.4\%) & 2 / 3 (66.7\%) \\
                                               & Gemini 2.5 Flash & \textbf{0 / 769 (0.0\%)} & \textbf{0 / 769 (0.0\%)} & - \\
        \rowcolor{gray!10}
                             \cellcolor{white} & Llama 3.3 70B    & 15 / 769 (2.0\%) & 2 / 769 (0.3\%) & 2 / 2 (100.0\%) \\
        \rowcolor{gray!10}
                             \cellcolor{white} & Qwen 3 32B       & 21 / 769 (2.7\%) & 10 / 769 (1.3\%) & 6 / 10 (60.0\%) \\
        \midrule
        \multirow[t]{5}{*}{\textbf{Tertiary}}  & GPT-5 Mini       & 6 / 5,052 (0.1\%) & 47 / 5,052 (0.9\%) & 20 / 47 (42.6\%) \\
                                               & GPT-5 Nano       & 231 / 5,052 (4.6\%) & 19 / 5,052 (0.4\%) & 12 / 19 (63.2\%) \\
                                               & Gemini 2.5 Flash & \textbf{0 / 5,052 (0.0\%)} & \textbf{0 / 5,052 (0.0\%)} & - \\
        \rowcolor{gray!10}
                             \cellcolor{white} & Llama 3.3 70B    & 236 / 5,052 (4.7\%) & 37 / 5,052 (0.7\%) & 22 / 37 (59.5\%) \\
        \rowcolor{gray!10}
                             \cellcolor{white} & Qwen 3 32B       & 144 / 5,052 (2.9\%) & 74 / 5,052 (1.5\%) & \textbf{29 / 74 (39.2\%)} \\
        \bottomrule
        \end{tabular}
    }
    \end{center}
\end{table}

%% file: algorithm/synthesize_algorithm.tex
\begin{algorithm}[t]
\caption{Synthetic data generation process}
\label{alg:synthetic-data}
\small
\KwIn{
Hospital $H_i$ (size $n_h$); Department $D_{ij}$ ($n_d \in [m_d, M_d]$); Physician $PH_{ijk}$ ($n_{ph} \in [m_{ph}, M_{ph}]$); Patient $PT_{ijkl}$ ($n_{pt} \in [m_{pt}, M_{pt}]$)
}
\For{$i \gets 1$ \KwTo $n_h$}{
    Initialize $H_i$ (\eg time system)\\
    $n_d \gets \text{random}(m_d, M_d)$\\
    \For{$j \gets 1$ \KwTo $n_d$}{
        $n_{ph} \gets \text{random}(m_{ph}, M_{ph})$\\
        \For{$k \gets 1$ \KwTo $n_{ph}$}{
            Generate physician profile $PH_{ijk}$\\
            $n_{pt} \gets \text{random}(m_{pt}, M_{pt})$\\
            \For{$l \gets 1$ \KwTo $n_{pt}$}{
                Generate patient profile $PT_{ijkl}$
            }
        }
    }
}
\end{algorithm}

%% file: appd_tables/synthesized_data.tex
\begin{table*}[!ht]
    \caption{Synthesized data specifications for hospital, physician, and patient entities. Italicized variables indicate parameters that are randomly assigned or generated during data synthesis.}
    \label{tab:synthesized data}
    \begin{center}
        \resizebox{1.0\linewidth}{!}{
            \begin{tabular}{p{4cm} p{12cm} p{7.5cm}}
            \toprule
            \textbf{Entity Category} & \textbf{Variables (YAML Parameter)} & \textbf{Variable Examples of YAML Format} \\
            \midrule
            \textbf{Hospital} &
                \vspace{-\baselineskip}
                \begin{itemize}[leftmargin=5mm, itemsep=0pt, topsep=2pt, after=\vspace{-\baselineskip}]
                    \item Number of hospitals ($hospital\_n$)
                    \item Simulation start date ($start\_date$)
                    \item Simulation periods ($days$)
                    \item Time unit ($time\_unit$)
                    \item Hospital start hour ($start\_hour$)
                    \item Hospital end hour ($end\_hour$)
                    \item The number of departments per hospital ($departemnt\_per\_hospital$)
                    \item The number of physicians per department ($physician\_per\_department$)
                \end{itemize} &
                \vspace{-\baselineskip}
                \begin{itemize}[leftmargin=5mm, itemsep=0pt, topsep=2pt]
                    \item[] $hospital\_n: 10$
                    \item[] $start\_date:$
                    \begin{itemize}[leftmargin=5mm, itemsep=0pt, topsep=-4pt]
                        \item[] $min:$ 2025-03-17
                        \item[] $max:$ 2025-09-21
                    \end{itemize}
                    \item[] $days: 7$
                    \item[] $time\_unit: 0.25$
                    \item[] $start\_hour:$
                    \begin{itemize}[leftmargin=5mm, itemsep=0pt, topsep=-4pt]
                        \item[] $min: 9$
                        \item[] $max: 10$
                    \end{itemize}
                    \item[] $end\_hour:$
                    \begin{itemize}[leftmargin=5mm, itemsep=0pt, topsep=-4pt]
                        \item[] $min: 18$
                        \item[] $max: 19$
                    \end{itemize}
                    \item[] $department\_per\_hospital:$
                    \begin{itemize}[leftmargin=5mm, itemsep=0pt, topsep=-4pt]
                        \item[] $min: 7$
                        \item[] $max: 9$
                    \end{itemize}
                    \item[] $physician\_per\_department:$
                    \begin{itemize}[leftmargin=5mm, itemsep=0pt, topsep=-4pt]
                        \item[] $min: 2$
                        \item[] $max: 3$
                    \end{itemize}
                \end{itemize} \\
            \midrule
            \textbf{Physician} &
            \vspace{-\baselineskip}
                \begin{itemize}[leftmargin=5mm, itemsep=0pt, topsep=2pt, after=\vspace{-\baselineskip}]
                    \item \textit{Name}
                    \item \textit{Gender}
                    \item \textit{Birth date}
                    \item \textit{Telecom}
                    \item Working days ($working\_days$) 
                    \item Capacity per hour ($capacity\_per\_hour$)
                    \item Unavailable schedule probability ($busy\_schedule\_prob$)
                    \item Unavailable schedule ratio ($busy\_schedule\_ratio$)
                    \item Appointment ratio ($appointment\_ratio$)
                \end{itemize} &
                \vspace{-\baselineskip}
                \begin{itemize}[leftmargin=5mm, itemsep=0pt, topsep=2pt]
                    \item[] $working\_days:$
                    \begin{itemize}[leftmargin=5mm, itemsep=0pt, topsep=-4pt]
                        \item[] $min: 3$
                        \item[] $max: 4$
                    \end{itemize}
                    \item[] $capacity\_per\_hour:$
                    \begin{itemize}[leftmargin=5mm, itemsep=0pt, topsep=-4pt]
                        \item[] $min: 1$
                        \item[] $max: 4$
                    \end{itemize}
                    \item[] $busy\_schedule\_prob: 0$
                    \item[] $busy\_schedule\_ratio:$
                    \begin{itemize}[leftmargin=5mm, itemsep=0pt, topsep=-4pt]
                        \item[] $min: 0.4$
                        \item[] $max: 0.6$
                    \end{itemize}
                    \item[] $appointment\_ratio:$
                    \begin{itemize}[leftmargin=5mm, itemsep=0pt, topsep=-4pt]
                        \item[] $min: 0.2$
                        \item[] $max: 0.5$
                    \end{itemize}
                \end{itemize} \\
            \midrule
            \textbf{Patient} &
            \vspace{-\baselineskip}
                \begin{itemize}[leftmargin=5mm, itemsep=0pt, topsep=2pt, after=\vspace{-\baselineskip}]
                    \item \textit{Name}
                    \item \textit{Gender}
                    \item \textit{Birth date}
                    \item \textit{Telecom}
                    \item \textit{Identifier}
                    \item \textit{Address}
                    \item Preference ($preference$) 
                    \item Symptom level ($symtom$)
                \end{itemize} &
                \vspace{-\baselineskip}
                \begin{itemize}[leftmargin=5mm, itemsep=0pt, topsep=2pt]
                    \item[] $preference:$
                    \begin{itemize}[leftmargin=5mm, itemsep=0pt, topsep=-4pt]
                        \item[] $type: [asap, physician, date]$
                        \item[] $probs: [0.4, 0.4, 0.2]$
                    \end{itemize}
                    \item[] $symptom:$
                    \begin{itemize}[leftmargin=5mm, itemsep=0pt, topsep=-4pt]
                        \item[] $type: [without\_history, with\_history]$
                        \item[] $probs: [0.7, 0.3]$
                    \end{itemize}
                \end{itemize} \\
            \bottomrule
            \end{tabular}
        }
    \end{center}
\end{table*}

%% file: prompts/crawling_prompt.tex
\begin{tcolorbox}[
    title={LLM prompt for post-processing crawled data},
    width=1\linewidth,
    colback=white,
    colframe=black,
    fonttitle=\bfseries,
    coltitle=white,
    breakable 
]
\textcolor{lightgray}{[\texttt{System Prompt}]} \\
You are an assistant that summarizes the symptoms of each disease based on the content crawled from disease-related web pages.\\

\textcolor{lightgray}{[\texttt{User Prompt}]} \\
Here is a disease and the content crawled from a web page containing information about this disease.\\

Disease: \texttt{\{DISEASE\}}\\
Web page crawling result:\\
\texttt{\{WEBPAGE\}}\\

Please summarize the possible symptoms that may occur when this disease develops, following the format below.
If no symptoms can be identified, respond with "No symptoms".\\

Answer format examples:\\
\{Disease Name\}\\
* \{symptom 1\}\\
* \{symptom 2\}\\
* \{symptom 3\}\\
...
\end{tcolorbox}

%% file: appd_tables/disease_labeling.tex
\begin{table*}[!ht]
\caption{Golden labels for disease--symptom pairs, listing the responsible departments, the number of diseases per department, and representative examples. Italicized disease names indicate cases in which treatment is possible across multiple departments.}
\label{tab:disease labeling results}
\centering
\resizebox{1.0\linewidth}{!}{
    \begin{tabular}{p{4.5cm} p{2.5cm} p{10cm}}
    \toprule
    \textbf{Department} & \textbf{Disease Count} & \textbf{Disease Examples} \\
    \midrule
    \textbf{Gastroenterology} & 35 & Escherichia coli (E. coli) O157,  \textit{Gastroenteritis}, Coeliac disease, Stomach ulcer, Norovirus, Inflammatory bowel disease (IBD), Hepatitis B, Food poisoning, etc. \\
    \midrule
    \textbf{Cardiology} & 19 & Abdominal aortic aneurysm, Angina, Arterial thrombosis, Atrial fibrillation, Heart attack, Heart block, Supraventricular tachycardia, etc. \\
    \midrule
    \textbf{Pulmonology} & 27 & Sepsis, Asbestosis, \textit{Asthma}, Breathlessness, Bronchiectasis, Catarrh, \textit{Chest infection}, \textit{Cystic fibrosis}, \textit{Carcinoid syndrome and carcinoid tumours}, etc.\\
    \midrule
    \textbf{Endocrinology/Metabolism} & 15 & Addison’s disease, \textit{Cystic fibrosis}, Dehydration, \textit{Gastroenteritis}, High blood sugar, Obesity, Osteoporosis, Overactive thyroid, Diabetes (Type 1, Type 2), etc. \\
    \midrule
    \textbf{Nephrology} & 5 & Chronic kidney disease, \textit{Kidney cancer}, \textit{Kidney infection}, Kidney stones, Urinary tract infection (UTI) \\
    \midrule
    \textbf{Hematology/Oncology} & 49 & Acute lymphoblastic leukaemia, Aplastic anaemia, \textit{Carcinoid syndrome and carcinoid tumours}, Ewing sarcoma, \textit{Kidney cancer}, Mesothelioma, etc. \\
    \midrule
    \textbf{Infectious diseases} & 42 & Clostridium difficile, \textit{Chest infection}, \textit{Croup}, Ebola virus disease, \textit{Kidney infection}, Threadworms, Thrush, Tick bites, Strep A, etc. \\
    \midrule
    \textbf{Allergy} & 18 & \textit{Asthma}, Food allergy, Hay fever, Hives, Allergic rhinitis, Allergies, etc. \\
    \midrule
    \textbf{Rheumatology} & 18 & \textit{Ankylosing spondylitis}, \textit{Arthritis}, \textit{Costochondritis}, \textit{Fibromyalgia}, \textit{Gout}, \textit{Lupus}, etc. \\
    \bottomrule
    \end{tabular}
}
\end{table*}

%% file: appd_figures/synthetic_data_example.tex
\begin{tcolorbox}[
    title={Example of synthesized data for the simulation},
    width=1\linewidth,
    colback=white,
    colframe=blue!50!black,
    fonttitle=\bfseries,
    coltitle=white,
    breakable 
]
\begin{footnotesize}
\begin{verbatim}
{
    "metadata": {
        "hospital_name": "hospital_00",
            "start_date": "2025-04-17",
            "end_date": "2025-04-18",
            "days": 2,
            "department_num": 1,
            "doctor_num": 1,
            "time": {"start_hour": 9.0, "end_hour": 18.0, "time_unit": 0.25}
    },
    "department": {
        "endocrinology/metabolism": {"code": "IMEND", "doctor": ["Dr. Benedict Tomerlin"]}
    },
    "doctor": {
        "Dr. Benedict Tomerlin": {
            "department": "endocrinology/metabolism",
            "specialty": {"name": "Osteoporosis and Metabolic Bone Disease", "code": "IMEND-2"},
            "schedule": {"2025-04-17": [], "2025-04-18": [[9.0, 18.0]]},
            "capacity_per_hour": 4,
            "capacity": 36,
            "gender": "female",
            "telecom": [{"system": "phone", "value": "+82844559851970", "use": "work"}],
            "birthDate": "1961-03-21"
        }
    },
    "patient": [{
        "patient": "Reynaldo Verlotte",
        "gender": "female",
        "telecom": [{"system": "phone", "value": "+8275129708711", "use": "work"}],
        "birthDate": "1982-09-21",
        "identifier": [{"value": "820921-1133985", "use": "official"}],
        "address": [{"type": "postal", "text": "0, Hoedong-ro, Yecheon-gun, Gunsan-si", "use": "home"}],
        "constraint": {
            "preference": ["physician", "asap"],
            "attending_physician": "Dr. Benedict Tomerlin",
            "valid_from": "N/A",
            "symptom_level": "without_history",
            "symptom": {
                "disease": "Dehydration",
                "department": ["endocrinology/metabolism"],
                "symptom": ["Peeing less often than usual", "Headache"]
            }
        }
    }]
}
\end{verbatim}
\end{footnotesize}
\end{tcolorbox}

%% file: appd_figures/fhir_resources.tex
\begin{tcolorbox}[
    title={Example of FHIR resources},
    width=1\linewidth,
    colback=white,
    colframe=blue!50!black,
    fonttitle=\bfseries,
    coltitle=white,
    breakable 
]
\begin{footnotesize}
\begin{verbatim}
======================================= Practitioner =======================================
{
    "resourceType": "Practitioner",
    "id": "hospital01-imrhe-Dr.LincolnBendzus",
    "active": true,
    "name": [{"family": "Bendzus", "given": ["Lincoln"], "prefix": ["Dr."]}],
    "telecom": [{"system": "phone", "value": "+825959635756519", "use": "work"}],
    "gender": "male",
    "birthDate": "1960-03-30"
}
===================================== PractitionerRole =====================================
{
    "resourceType": "PractitionerRole",
    "id": "hospital01-imall-Dr.ThurmanModdejonge-role",
    "active": true,
    "practitioner": {"reference": "Practitioner/hospital01-imall-Dr.ThurmanModdejonge"},
    "specialty": [
        {"coding": [{"code": "IMALL-2", "display": "Food and Drug Allergy"}], "text": "allergy"}
    ],
    "characteristic": [
        {"coding": [{"code": "capacity_per_hour", "display": "4"}], "text": "capacity_per_hour"}, 
        {"coding": [{"code": "capacity","display": "192"}], "text": "capacity"}
    ]
}
========================================= Schedule =========================================
{
    "resourceType": "Schedule",
    "id": "hospital01-imrhe-Dr.LincolnBendzus-schedule",
    "active": true,
    "actor": [{"reference": "Practitioner/hospital01-imrhe-Dr.LincolnBendzus"}],
    "planningHorizon": {"start": "2025-03-22T10:00:00+09:00", "end": "2025-03-28T18:00:00+09:00"}
}
=========================================== Slot ===========================================
{
  "resourceType": "Slot",
  "id": "hospital01-imrhe-Dr.LincolnBendzus-20250325-slot10",
  "schedule": {"reference": "Schedule/hospital01-imrhe-Dr.LincolnBendzus-schedule"},
  "status": "busy",
  "start": "2025-03-25T12:30:00+09:00",
  "end": "2025-03-25T12:45:00+09:00"
}
========================================== Patient =========================================
{
    "resourceType": "Patient",
    "id": "hospital02-iminf-EarnestMetoxen",
    "identifier": [{"use": "official", "value": "961010-7363710"}],
    "active": true,
    "name": [{"family": "Metoxen", "given": ["Earnest"]}],
    "telecom": [{"system": "phone", "value": "+82693618062105", "use": "mobile"}],
    "gender": "male",
    "birthDate": "1996-10-10",
    "address": [{"use": "home", "type": "postal", "text": "4, Dongbang, Geumsan-gun, Hanam-si"}]
}
======================================= Appointment ========================================
{
    "resourceType": "Appointment",
    "id": "hospital02-iminf-Dr.AldenGaestel-20250918-appn0-0",
    "status": "booked",
    "start": "2025-09-18T10:00:00+09:00",
    "end": "2025-09-18T10:15:00+09:00",
    "slot": [{"reference": "Slot/hospital02-iminf-Dr.AldenGaestel-20250918-slot0"}],
    "participant": [
        {"actor": {
            "reference": "Practitioner/hospital02-iminf-Dr.AldenGaestel", 
            "display": "Dr. Alden Gaestel"
        }, "status": "accepted"}, 
        {"actor": {
            "reference": "Patient/hospital02-iminf-EarnestMetoxen", 
            "display": "Earnest Metoxen"
        }, "status": "accepted"}
    ]
}
\end{verbatim}
\end{footnotesize}
\end{tcolorbox}

%% file: appd_tables/rubric_table.tex
\begin{table*}[ht!]
    \caption{Defined error codes and suspected causes categorized by task type. The suspected causes list possible reasons for each error code.}
    \label{tab:rubric table}
    \centering
    \resizebox{1\linewidth}{!}{
        \begin{tabular}{p{4cm} p{4.5cm} p{13.5cm}}
        \toprule
        \textbf{Tasks} & \textbf{Error Codes} & \textbf{Suspected Causes} \\
        \midrule
        \textbf{Common} & Incorrect Format (IF) &
        \vspace{-\baselineskip}
        \begin{itemize}[leftmargin=5mm, itemsep=0pt, topsep=2pt, after=\vspace{-\baselineskip}]
            \item There is an issue with the output format.
        \end{itemize} 
        \\
        \cmidrule{2-3}
                                & Incomplete Simulation (IS) &
        \vspace{-\baselineskip}
        \begin{itemize}[leftmargin=5mm, itemsep=0pt, topsep=2pt, after=\vspace{-\baselineskip}]
            \item The patient failed to accurately convey the provided information during the conversation with the administrative staff.
            \item The staff agent failed to complete all required actions to fulfill the patient’s request.
        \end{itemize}
        \\
        \midrule
        \textbf{Intake}         & Incorrect Department (ID) & 
        \vspace{-\baselineskip}
        \begin{itemize}[leftmargin=5mm, itemsep=0pt, topsep=2pt, after=\vspace{-\baselineskip}]
            \item The department finally assigned to the patient was incorrect.
        \end{itemize}
        \\
        \cmidrule{2-3}
                                & Incorrect Patient Information (IPI) &
        \vspace{-\baselineskip}
        \begin{itemize}[leftmargin=5mm, itemsep=0pt, topsep=2pt, after=\vspace{-\baselineskip}]
            \item The information extracted from the conversation did not match the information actually provided by the patient.
        \end{itemize}                
        \\
        \cmidrule{2-3}
                                & Incorrect Department \& Patient Information (IDPI) & 
        \vspace{-\baselineskip}
        \begin{itemize}[leftmargin=5mm, itemsep=0pt, topsep=2pt, after=\vspace{-\baselineskip}]
            \item The department assigned to the patient was incorrect, and the information extracted from the conversation did not match the patient’s actual information.
        \end{itemize}                
        \\
        \midrule
        \multirow[t]{8}{*}{\textbf{Scheduling}} & Physician Conflict (PC) & 
        \vspace{-\baselineskip}
        \begin{itemize}[leftmargin=5mm, itemsep=0pt, topsep=2pt, after=\vspace{-\baselineskip}]
            \item More than one doctor has been assigned.
        \end{itemize}
        \\
        \cmidrule{2-3}
                                & Invalid Schedule (IVS) &
        \vspace{-\baselineskip}
        \begin{itemize}[leftmargin=5mm, itemsep=0pt, topsep=5pt, after=\vspace{-\baselineskip}]
            \item The scheduling result may fall outside the hospital's operating hours.
            \item The scheduling result may be in the past relative to the current time.
            \item The scheduling result may not be a valid date.
            \item The assigned doctor may not belong to the department the patient should visit.
        \end{itemize}
        \\
        \cmidrule{2-3}
                                & Wrong Duration (WD) &
        \vspace{-\baselineskip}
        \begin{itemize}[leftmargin=5mm, itemsep=0pt, topsep=5pt, after=\vspace{-\baselineskip}]
            \item The patient's schedule does not match the consultation duration required by the doctor.
        \end{itemize}
        \\
        \cmidrule{2-3}
                                & Time Conflict (TC) &
        \vspace{-\baselineskip}
        \begin{itemize}[leftmargin=5mm, itemsep=0pt, topsep=5pt, after=\vspace{-\baselineskip}]
            \item The scheduling result overlaps with the doctor's existing schedule.
        \end{itemize}
        \\
        \cmidrule{2-3}
                                & Incorrect Physician (IP) & 
        \vspace{-\baselineskip}
        \begin{itemize}[leftmargin=5mm, itemsep=0pt, topsep=5pt, after=\vspace{-\baselineskip}]
            \item A different doctor was assigned even though the patient requested a specific doctor.
        \end{itemize}
        \\
        \cmidrule{2-3}
                                & Incorrect Date (IDT) &
        \vspace{-\baselineskip}
        \begin{itemize}[leftmargin=5mm, itemsep=0pt, topsep=5pt, after=\vspace{-\baselineskip}]
            \item The patient is available only after a specific date, but the appointment is scheduled before that date.
        \end{itemize}
        \\
        \cmidrule{2-3}
                                & Not the Earliest Time (NET) &
        \vspace{-\baselineskip}
        \begin{itemize}[leftmargin=5mm, itemsep=0pt, topsep=5pt, after=\vspace{-\baselineskip}]
            \item The patient wants the earliest possible appointment in the department, but the assigned time is not the earliest available based on the current time.
            \item When scheduling, it is possible to assign an earlier date or time.
        \end{itemize}
        \\
        \midrule
        \textbf{Event}         & Fail to Identify (FI) & 
        \vspace{-\baselineskip}
        \begin{itemize}[leftmargin=5mm, itemsep=0pt, topsep=2pt, after=\vspace{-\baselineskip}]
            \item The existing appointment schedule could not be identified from the patient’s request.
        \end{itemize}
        \\
        \bottomrule
        \end{tabular}
    }
\end{table*}

%% file: tables/token_cost.tex
\begin{table*}[t!]
    \caption{Detailed breakdown of token consumption and operational costs per simulation run across different LLMs and administrative tasks. \textit{Scheduling (T)} uses tool-calling, while \textit{Scheduling (R)} uses LLM reasoning only.}
    \label{tab:token_cost}
    \small
    \begin{center}
    \resizebox{1\linewidth}{!}{
        \begin{tabular}{llrrrrrrrr}
        \toprule
        \multirow[c]{3}{*}{\textbf{Model}} & \multirow[c]{3}{*}{\textbf{Task}} & \multicolumn{4}{c}{\textbf{Patient Agent}} & \multicolumn{4}{c}{\textbf{Staff Agent}} \\
        \cmidrule(lr{2pt}{2pt}){3-6} \cmidrule(lr{2pt}{2pt}){7-10}
        & & \multirow[c]{2}{*}{\textbf{Cost (\$)}} & \multicolumn{3}{c}{\textbf{Tokens}} & \multirow[c]{2}{*}{\textbf{Cost (\$)}} & \multicolumn{3}{c}{\textbf{Tokens}} \\
        \cmidrule(lr{2pt}{2pt}){4-6} \cmidrule(lr{2pt}{2pt}){8-10}
        & & & \makecell[c]{\textbf{Input}} & \makecell[c]{\textbf{Reasoning}} & \makecell[c]{\textbf{Output}} & & \makecell[c]{\textbf{Input}} & \makecell[c]{\textbf{Reasoning}} & \makecell[c]{\textbf{Output}} \\
        \midrule
        \multirow[t]{3}{*}{\textbf{GPT-5 Mini}}        & Intake         & 0.003840 & 7063.8 & 422.4 & 614.8 & 0.006246 & 4867.1 & 1021.4 & 1493.2 \\
                                                       & Scheduling (T) & 0.001008 & 755.0  & 186.9 & 222.5 & 0.001406 & 1636.9 & 231.7  & 266.6 \\
                                                       & Scheduling (R) & 0.000958 & 740.1 & 175.4 & 211.4 & 0.001935 & 2209.7 & 314.9 & 376.3 \\
        \midrule
        \multirow[t]{3}{*}{\textbf{GPT-5 Nano}}        & Intake         & 0.000947 & 5434.6 & 755.2 & 932.4 & 0.001294 & 3760.4 & 1260.8 & 1505.4 \\
                                                       & Scheduling (T) & 0.000237 & 726.1 & 233.0 & 267.7 & 0.000659 & 1583.9 & 707.8 & 740.9 \\
                                                       & Scheduling (R) & 0.000242 & 698.9 & 241.9 & 276.0 & 0.000631 & 1973.7 & 636.2 & 695.7 \\
        \midrule
        \multirow[t]{3}{*}{\textbf{Gemini 2.5 Flash}}  & Intake         & 0.004444 & 6640.2 & 833.2 & 147.8 & 0.006072 & 4448.1 & 1587.7 & 307.2 \\
                                                       & Scheduling (T) & 0.001878 & 762.7 & 637.7 & 21.9  & 0.000882 & 1743.6 & 58.7 & 84.9 \\
                                                       & Scheduling (R) & 0.001635 & 720.2 & 546.9 & 20.7 & 0.008952 & 1979.9 & 3258.8 & 84.5 \\
        \bottomrule
        \end{tabular}
    }
    \end{center}
\end{table*}

%% file: appd_tables/human_eval.tex
\begin{table}[t!]
    \caption{Summary of the human evaluation. The \textit{Arena} and \textit{Score} columns report pairwise win counts and the average 5-point ratings across four evaluation criteria, respectively.}
    \label{tab:human eval}
    \centering
    \small
    \begin{tabular}{lrr}
    \toprule
    \textbf{Models} & \textbf{Arena} & \textbf{Score} \\
    \midrule
    \textbf{GPT-5 Mini}        & 42 & 3.33$\pm$1.31 \\
    \textbf{GPT-5 Nano}        & 80 & 4.06$\pm$1.12 \\
    \textbf{Gemini 2.5 Flash}  & \textbf{89} & \textbf{4.11$\pm$1.20}  \\
    \bottomrule
    \end{tabular}
\end{table}

%% file: appd_tables/error_analysis_schedule_R.tex
\begin{table*}[t!]
    \caption{Error distribution in the LLM reasoning-based scheduling task by hospital level, model, and error code. The prefixes (\eg \textit{C3-1}, \textit{C3-2}) denote the evaluation criteria and the order in which they are evaluated, where criterion \textit{C3-7} can yield one of two error codes, and \textit{C4-1} applies only to rescheduling and cancellation (Appendix~\ref{appendix:experimental details:rubrics}). Shaded cells indicate open-source models. Bold and underlined values denote the best and second-best performance, respectively. \textit{Abbreviations: IS = Incomplete Simulation, IF = Incorrect Format, PC = Physician Conflict, IVS = Invalid Schedule, WD = Wrong Duration, TC = Time Conflict, IP = Incorrect Physician, IDT = Incorrect Date, NET = Not the Earliest Time, FI = Fail to Identify.}}
    \label{tab:error_analysis_scheduling_R}
    \begin{center}
        \resizebox{1\linewidth}{!}{
            \begin{tabular}{llr||rrrrrrrrr|r}
            \toprule
            \multirow[c]{2}{*}{\textbf{Hospital}} & \multirow[c]{2}{*}{\textbf{Model}} & \multicolumn{11}{c}{\textbf{Scheduling (R)}} \\
            \cmidrule{3-13}
            & & \makecell[c]{\textbf{Errors}} & \makecell[c]{\textbf{C3-1. IS}} & \makecell[c]{\textbf{C3-2. IF}} & \makecell[c]{\textbf{C3-3. PC}} & \makecell[c]{\textbf{C3-4. IVS}} & \makecell[c]{\textbf{C3-5. WD}} & \makecell[c]{\textbf{C3-6. TC}} & \makecell[c]{\textbf{C3-7. IP}} & \makecell[c]{\textbf{C3-7. IDT}} & \makecell[c]{\textbf{C3-8. NET}} & \makecell[c]{\textbf{C4-1. FI}} \\
            \midrule
            \multirow[t]{5}{*}{\textbf{Primary}}   & GPT-5 Mini       & \textbf{6 / 586 (1.0\%)}     & - & - & - & - & - & - & - & - & 6 (100.0\%) & - \\
                                                   & GPT-5 Nano       & 284 / 651 (43.6\%)  & - & - & - & - & 22 (7.8\%) & 13 (4.6\%) & - & 2 (0.7\%) & 247 (87.0\%) & - \\
                                                   & Gemini 2.5 Flash & 402 / 635 (63.3\%)  & - & - & - & 1 (0.3\%) & - & - & - & 2 (0.5\%) & 398 (99.0\%) & 1 (0.3\%) \\
            \rowcolor{gray!10}
                                 \cellcolor{white} & Llama 3.3 70B    & 353 / 577 (61.2\%)  & - & 11 (3.1\%) & - & 21 (6.0\%) & - & 26 (7.4\%) & - & 1 (0.3\%) & 266 (75.4\%) & 28 (7.9\%) \\
            \rowcolor{gray!10}
                                 \cellcolor{white} & Qwen 3 32B       & \underline{242 / 590 (41.0\%)}  & - & 1 (0.4\%) & - & - & 4 (1.7\%) & 7 (2.9\%) & - & - & 226 (93.4\%) & 4 (1.7\%) \\
            \midrule
            \multirow[t]{5}{*}{\textbf{Secondary}} & GPT-5 Mini       & \textbf{32 / 906 (3.5\%)}    & - & 1 (3.1\%) & - & - & - & - & - & - & 31 (96.9\%) & - \\
                                                   & GPT-5 Nano       & 316 / 932 (33.9\%)  & - & 1 (0.3\%) & - & - & 21 (6.7\%) & 81 (25.6\%) & 3 (1.0\%) & 1 (0.3\%) & 209 (66.1\%) & - \\
                                                   & Gemini 2.5 Flash & 306 / 913 (33.5\%)  & - & - & - & - & - & 2 (0.7\%) & 1 (0.3\%) & - & 302 (98.7\%) & 1 (0.3\%) \\
            \rowcolor{gray!10}
                                 \cellcolor{white} & Llama 3.3 70B    & 603 / 893 (67.5\%)  & - & - & - & 5 (0.8\%) & - & 32 (5.3\%) & - & 6 (1.0\%) & 519 (86.1\%) & 41 (8.8\%) \\
            \rowcolor{gray!10}
                                 \cellcolor{white} & Qwen 3 32B       & \underline{280 / 935 (30.0\%)}  & - & 5 (1.8\%) & - & - & 3 (1.1\%) & 44 (15.7\%) & 2 (0.7\%) & - & 218 (77.9\%) & 8 (2.9\%) \\
            \midrule
            \multirow[t]{5}{*}{\textbf{Tertiary}}  & GPT-5 Mini       & \textbf{2,044 / 8,159 (25.1\%)}     & - & - & - & - & 317 (15.5\%) & 45 (2.2\%) & - & - & 1,682 (82.3\%) & - \\
                                                   & GPT-5 Nano       & 12,550 / 15,197 (82.6\%) & - & 93 (0.7\%) & - & 7 (0.1\%) & 691 (5.5\%) & 1,523 (12.1\%) & 73 (0.6\%) & 16 (0.1\%) & 10,145 (80.8\%) & 2 (0.0\%) \\
                                                   & Gemini 2.5 Flash & \underline{7,669 / 12,264 (62.5\%)}    & - & - & - & 31 (0.4\%) & 182 (2.4\%) & 161 (2.1\%) & 1 (0.0\%) & 250 (3.3\%) & 7,043 (91.8\%) & 1 (0.0\%) \\
            \rowcolor{gray!10}
                                 \cellcolor{white} & Llama 3.3 70B    & 6,089 / 6,997 (87.0\%)  & - & 4 (0.1\%) & - & 22 (0.4\%) & 1,077 (17.7\%) & 303 (5.0\%) & - & 35 (0.6\%) & 3,749 (61.6\%) & 899 (14.8\%) \\
            \rowcolor{gray!10}
                                 \cellcolor{white} & Qwen 3 32B & 9,972 / 12,994 (76.7\%)  & - & 63 (0.6\%) & - & - & 59 (0.6\%) & 1,025 (10.3\%) & 40 (0.4\%) & - & 8,671 (87.0\%) & 114 (1.1\%) \\
            \bottomrule
            \end{tabular}
        }
    \end{center}
\end{table*}

%% file: appd_tables/persona_desc.tex
\begin{table*}[ht!]
    \caption{Patient personas used in the outpatient intake simulation, with detailed behavioral and linguistic descriptions.}
    \label{tab:persona desc}
    \centering
    \resizebox{1\linewidth}{!}{
        \begin{tabular}{p{4cm} p{2cm} p{15cm}}
        \toprule
        \textbf{Persona Type} & \textbf{Level} & \textbf{Description} \\
        \midrule
        \textbf{Personality} & Neutral &
        \vspace{-\baselineskip}
        \begin{enumerate}[leftmargin=5mm, itemsep=0pt, topsep=2pt, after=\vspace{-\baselineskip}]
            \item Provide concise, direct answers focused on the question, without extra details.
            \item Respond in a neutral tone without any noticeable emotion or personality.
        \end{enumerate} 
        \\
        \midrule
        \textbf{Language Proficiency} & Intermediate (CEFR B) &
        Act as a patient with intermediate English proficiency (CEFR B). You must:
        \begin{enumerate}[leftmargin=5mm, itemsep=0pt]
            \item Speaking: Use common vocabulary and form connected, coherent sentences with occasional minor grammar errors. Discuss familiar topics confidently but struggle with abstract or technical subjects. Avoid highly specialized or abstract words.
            \item Understanding: Can understand the main ideas of everyday conversations. Need clarification or simpler explanations for abstract, technical, or complex information. Words within your level: \{understand\_words\}. Words beyond your level: \{misunderstand\_words\}.
            \item Medical Terms: Use and understand common medical terms related to general health. Cannot use or understand advanced or specialized medical terms and require these to be explained in simple language. Below are examples of words within and beyond your level. You cannot understand words more complex than the examples provided within your level. Words within your level: \{understand\_med\_words\}. Words beyond your level: \{misunderstand\_med\_words\}.
        \end{enumerate}
        IMPORTANT: If a question contains advanced terms beyond your level, ask for simpler explanation (\eg `I don’t get it' or `What do you mean?'). Keep asking until the question is clear enough for you to answer. \\
        \midrule
        \textbf{Confusion} & Normal &
        Clearly understand the question according to the CEFR level, and naturally reflect your background
        and personality in your responses. \\
        \midrule
        \multirow[t]{2}{*}{\textbf{Recall}} & No history &
        \vspace{-\baselineskip}
        \begin{enumerate}[leftmargin=5mm, itemsep=0pt, topsep=2pt, after=\vspace{-\baselineskip}]
            \item No previous diagnoses or surgeries are documented.
            \item No chronic conditions, regular medications, or relevant family medical history are reported.
        \end{enumerate} \\
        \cmidrule{2-3}
         & High &
        \vspace{-\baselineskip}
        \begin{enumerate}[leftmargin=5mm, itemsep=0pt, topsep=5pt, after=\vspace{-\baselineskip}]
            \item Accurately remember all health-related information, including past conditions, current medications, and other documented details.
            \item Do not forget or confuse medical information.
            \item Consistently ensure that recalled details match documented records.
        \end{enumerate} \\
        \bottomrule
        \end{tabular}
    }
\end{table*}

%% file: appd_tables/preference_desc.tex
\begin{table*}[ht!]
    \caption{Preference descriptions used in the scheduling task for the patient agent.}
    \label{tab:preference_desc}
    \centering
    \resizebox{1\linewidth}{!}{
        \begin{tabular}{ll}
        \toprule
        \textbf{Preference Type} & \textbf{Description} \\
        \midrule
        \textbf{ASAP} & You want the earliest available doctor in the department for the outpatient visit.\\
        \textbf{Physician} & You have a preferred doctor for the outpatient visit.\\
        \textbf{Date} & You want the earliest available doctor in the department for the outpatient visit, starting from **\{date\}**.\\
        \bottomrule
        \end{tabular}
    }
\end{table*}

%% file: prompts/intake_patient_system_prompt.tex
\begin{tcolorbox}[
    title={Patient agent prompt template for intake role-playing},
    width=1\linewidth,
    colback=white,
    colframe=black,
    fonttitle=\bfseries,
    coltitle=white,
    breakable 
]
Imagine you are a patient experiencing physical or emotional health concerns.
You are visiting the hospital for the first time and need to consult with the administration office to determine the appropriate department for your symptoms.
Your task is to role-play this patient during an initial consultation with the hospital administration staff, providing the necessary information to guide the department selection.\\

Patient Background Information:
\begin{itemize}[leftmargin=7mm, itemsep=0pt, topsep=1pt, parsep=1pt, partopsep=1pt]
    \item Personal Information:
        \begin{itemize}[leftmargin=7mm, itemsep=0pt, topsep=1pt, parsep=1pt, partopsep=1pt]
            \item Name: \texttt{\{name\}}
            \item Gender: \texttt{\{gender\}}
            \item Phone Number: \texttt{\{telecom\}}
            \item Birth Date: \texttt{\{birth\_date\}}
            \item Personal ID: \texttt{\{personal\_id\}}
            \item Address: \texttt{\{address\}}
        \end{itemize}
    \item Previous Medical History:
        \begin{itemize}[leftmargin=7mm, itemsep=0pt, topsep=1pt, parsep=1pt, partopsep=1pt]
            \item Allergies: \texttt{\{allergies\}}
            \item Family medical history: \texttt{\{family\_medical\_history\}}
            \item Relevant medical history: \texttt{\{medical\_history\}}
        \end{itemize}
    \item Current disease and symptom:
        \begin{itemize}[leftmargin=7mm, itemsep=0pt, topsep=1pt, parsep=1pt, partopsep=1pt]
            \item Disease: \texttt{\{diagnosis\}}
            \item Symptom: \texttt{\{chief\_complaint\}}
        \end{itemize}
    \item Note:
        \begin{itemize}[leftmargin=7mm, itemsep=0pt, topsep=1pt, parsep=1pt, partopsep=1pt]
            \item You will be asked about your current symptoms and personal information. Describe symptoms that may be related to {department}, and engage in a conversation with the administration office staff based on the provided patient background information. And don’t mention any personal information until the administrative staff asks for it first.
            \item Keep in mind that although your symptoms may appear linked to a specific department, you do not yet know which department you should visit. Also, do not directly ask which department you should visit.
            \item The final decision will be made entirely by the administration office based on the previously diagnosed disease, relevant medical history, and symptoms you report. If the `Disease` field contains a specific illness and the `Relevant medical history` includes the same diagnosis, you are considered to be visiting the hospital for that illness. Therefore, if the administration office staff ask about your previous diagnosis history, you must respond based on it.
            \item When describing your condition, let your responses naturally reflect the persona information below, including the specified personality, language proficiency, medical history recall ability, and level of dazedness.
        \end{itemize}
\end{itemize}
\vspace{1em}
Persona:
\begin{itemize}[leftmargin=7mm, itemsep=0pt, topsep=1pt, parsep=1pt, partopsep=1pt]
    \item Personality: 
    \begin{itemize}[leftmargin=7mm, itemsep=0pt, topsep=1pt, parsep=1pt, partopsep=1pt]
        \item Personality: \texttt{\{personality\}}
        \item Language Proficiency: \texttt{\{lang\_proficiency\}}
        \item Medical History Recall Ability: \texttt{\{recall\}}
        \item Dazedness Level: \texttt{\{confusion\}}
    \end{itemize}
\end{itemize}
\vspace{1em}
In the consultation, simulate the patient described in the above profile, while the user plays the role of the administration office staff.
During the conversation, follow these guidelines:\\
\texttt{\{behavioral\_guideline\}} \\

You are now the patient.
Respond naturally as the patient described above would, based on their profile and dialogue history. 
Remember: \texttt{\{reminder\}}
You should answer within \texttt{\{sentence\_limit\}} sentences, keeping each sentence concise.
\end{tcolorbox}

%% file: prompts/intake_patient_guideline.tex
\begin{tcolorbox}[
    title={Patient agent behavioral guidelines for intake role-playing},
    width=1\linewidth,
    colback=white,
    colframe=black,
    fonttitle=\bfseries,
    coltitle=white,
    breakable 
]
\begin{enumerate}[leftmargin=7mm, itemsep=0pt, topsep=1pt, parsep=1pt, partopsep=1pt]
    \item Fully immerse yourself in the patient role, setting aside any awareness of being an AI model.
    \item Ensure responses stay consistent with the patient’s profile, current visit details, and prior conversation, allowing minor persona-based variations.
    \item Align responses with the patient’s language proficiency, using simpler terms or asking for rephrasing if any words exceed their level.
    \item Match the tone and style to the patient’s personality, reflecting it distinctly and naturally. Do not explicitly mention the personality.
    \item Minimize or exaggerate medical information, or even deny answers as appropriate, based on dazedness and personality.
    \item Prioritize dazedness over personality when dazedness is high, while maintaining language proficiency.
    \item Reflect the patient’s memory and dazedness level, potentially forgetting or confusing details.
    \item Keep responses realistic and natural. Avoid mechanical repetition and a robotic or exaggerated tone.
    \item Use informal, everyday language.
    \item Keep responses to 1–\texttt{\{sentence\_limit\}} concise sentences, each no longer than 20 words.
    \item Gradually reveal detailed information or experiences as the dialogue goes on. Avoid sharing all possible information without being asked.
    \item Respond only with what the patient would say, without describing physical actions or non-verbal cues.
    \item Do not directly reveal department assignment or final diagnosis, since the patient would not know this information.
    \item Do not directly ask which department you should visit. The final decision will be made entirely by the administration office based on the symptoms you report.
\end{enumerate} 
\end{tcolorbox}

%% file: prompts/intake_staff_system_prompt.tex
\begin{tcolorbox}[
    title={Administrative staff agent prompt template for intake role-playing},
    width=1\linewidth,
    colback=white,
    colframe=black,
    fonttitle=\bfseries,
    coltitle=white,
    breakable 
]
% \textcolor{lightgray}{[\texttt{System Prompt for intake role-playing}]} \\
You are playing the role of a kind and patient hospital administration staff.
Your job is to talk with the patient, listen to their main symptoms, and guide them step by step:
\begin{enumerate}[leftmargin=7mm, itemsep=0pt, topsep=1pt, parsep=1pt, partopsep=1pt]
    \item First decide if the patient should go to Internal Medicine or Surgery.
    \item Then guide the patient to the most suitable detailed department within that category.
\end{enumerate}
You may ask up to \texttt{\{total\_idx\}} questions before making your final decision.\\

Conversation guidelines:
\begin{enumerate}[leftmargin=7mm, itemsep=0pt, topsep=1pt, parsep=1pt, partopsep=1pt]
    \item You **must** ask about demographic information to the patient: Name, gender, phone number, personal ID, and address. Try to ask for all of the above information at once naturally, rather than separately.
    \item After obtaining the patient's demographic information, you **must** ask the patient about any previously diagnosed diseases.
    \item Focus on the patient’s main problem. Ask about:
        \begin{itemize}[leftmargin=7mm, itemsep=0pt, topsep=1pt, parsep=1pt, partopsep=1pt]
            \item Main symptom: when it started, how it feels, how long it lasts, and what makes it better or worse (use simple, everyday words), etc.
            \item Medical history: If the patient has diagnostic records or a diagnosis from a previous hospital, you should make the final decision on the department based on this information.\\
            Note: The patient has limited medical knowledge, so do not ask very detailed or technical questions about their symptoms.
        \end{itemize}
    \item Even if the patient does not have medical records or a diagnostic history from a previous hospital, you must not make a medical diagnosis yourself. Your purpose is to assign the most appropriate department for treatment, based on the previous hospital’s diagnosis if available, or on the patient’s symptoms if no such records exist.
    \item Avoid medical jargon. Use everyday words (e.g., say “yellowing of eyes” instead of “icterus”).
    \item Adjust your questions based on the patient’s answers. If unclear, gently rephrase.
    \item Show empathy and reassurance (e.g., “I understand that must be uncomfortable.”).
    \item Ask only one short and clear question at a time and keep your answers short (1–2 sentences per turn).
    \item Whenever you are able to determine the patient's department, you **must** use the following answer format, including the corresponding number from the options below. Three examples of the answer format:
    \begin{itemize}[leftmargin=7mm, itemsep=0pt, topsep=1pt, parsep=1pt, partopsep=1pt]
        \item `Answer: 1. orthopedics`
        \item `Answer: 4. neurology`
        \item `Answer: 3. oncology`
    \end{itemize}
\end{enumerate} 

\vspace{1em}
Current department options in the hospital:\\
\texttt{\{department\}}\\

This is round \texttt{\{curr\_idx\}}, you have \texttt{\{remain\_idx\}} rounds left.\\

% \textcolor{lightgray}{[\texttt{User Prompt after the role-playing}]} \\
% Please extract the patient's **name, gender, phone number, personal ID, address, and determined department**, and provide the output in the format shown in the examples below.\\

% \#\# Answer format examples:\\
% Example 1:\\
% \textasciigrave \textasciigrave \textasciigrave json\\
% \{"name": "Chuck Adamcik", "gender": "male", "phone\_number": "+82707813973", "personal\_id": "740925-6104539", "address": "93, Bomok 3-gil, Inje-gun, Gimpo-si", "department": "rheumatory"\}
% \textasciigrave \textasciigrave \textasciigrave \\

% Example 2:\\
% \textasciigrave \textasciigrave \textasciigrave json\\
% \{"name": "Garth Sumners", "gender": "female", "phone\_number": "+5832802426", "personal\_id": "856215-5992486", "address": "9, Jujung-ro 128beon-gil, Suyeong-gu, Yangsan-si", "department": "pulmonary"\}
% \textasciigrave \textasciigrave \textasciigrave \\

% Example 3 (if the demographic information is unavailable):\\
% \textasciigrave \textasciigrave \textasciigrave json \\
% \{"name": "none", "gender": "none", "phone\_number": "none", "personal\_id": "none", "address": "none", "department": "infectious diseases"\}
% \textasciigrave \textasciigrave \textasciigrave

\end{tcolorbox}

%% file: prompts/intake_staff_task_user_prompt.tex
\begin{tcolorbox}[
    title={Administrative staff agent prompt for post-dialogue processing in the intake task},
    width=1\linewidth,
    colback=white,
    colframe=black,
    fonttitle=\bfseries,
    coltitle=white,
    breakable 
]
Please extract the patient's **name, gender, phone number, personal ID, address, and determined department**, and provide the output in the format shown in the examples below.\\

\#\# Answer format examples:\\
Example 1:\\
\textasciigrave \textasciigrave \textasciigrave json\\
\{"name": "Chuck Adamcik", "gender": "male", "phone\_number": "+82707813973", "personal\_id": "740925-6104539", "address": "93, Bomok 3-gil, Inje-gun, Gimpo-si", "department": "rheumatory"\}
\textasciigrave \textasciigrave \textasciigrave \\

Example 2:\\
\textasciigrave \textasciigrave \textasciigrave json\\
\{"name": "Garth Sumners", "gender": "female", "phone\_number": "+5832802426", "personal\_id": "856215-5992486", "address": "9, Jujung-ro 128beon-gil, Suyeong-gu, Yangsan-si", "department": "pulmonary"\}
\textasciigrave \textasciigrave \textasciigrave \\

Example 3 (if the demographic information is unavailable):\\
\textasciigrave \textasciigrave \textasciigrave json \\
\{"name": "none", "gender": "none", "phone\_number": "none", "personal\_id": "none", "address": "none", "department": "infectious diseases"\}
\textasciigrave \textasciigrave \textasciigrave

\end{tcolorbox}

%% file: prompts/schedule_patient_system.tex
\begin{tcolorbox}[
    title={Patient agent system prompt template for the new appointment scheduling},
    width=1\linewidth,
    colback=white,
    colframe=black,
    fonttitle=\bfseries,
    coltitle=white,
    breakable 
]
Imagine you are a patient scheduling an appointment with hospital administrative staff.
You should naturally express your scheduling preferences to the staff member.

\vspace{1em}
\#\# Patient Scheduling Preference:
\begin{itemize}[leftmargin=7mm, itemsep=0pt, topsep=1pt, parsep=1pt, partopsep=1pt]
    \item Preference type: \texttt{\{preference\}}
    \item Preference explanation: \texttt{\{preference\_desc\}}
    \item Preferred doctor: \texttt{\{preferred\_doctor\}}
    \item Note:
    \begin{itemize}[leftmargin=7mm, itemsep=0pt, topsep=1pt, parsep=1pt, partopsep=1pt]
        \item ASAP preference case: If you have no additional conditions and want the earliest appointment in the department, state only that you want the earliest available doctor in that department.
        \item Doctor preference case: If you have a specific doctor in mind, you **must clearly and explicitly mention that doctor**, and the earliest available appointment with that doctor should be preferred.
        \item Date preference case: If you prefer an appointment after a specific date, you **must explicitly mention the exact date (year, month, and day)**, indicate that any available doctor on or after that date is acceptable, and must not express an ASAP or earliest-possible preference.
        \item If you do not have a preference for a specific doctor, state that any doctor is fine and naturally express your scheduling preference.
        \item Only mention a date if you explicitly prefer an appointment after a specific date; otherwise, do not mention any dates.
    \end{itemize}
\end{itemize}

\vspace{1em}
\#\# Persona:
\begin{itemize}[leftmargin=7mm, itemsep=0pt, topsep=1pt, parsep=1pt, partopsep=1pt]
    \item Personality: \texttt{\{personality\}}
\end{itemize}

\vspace{1em}
\#\# Guidelines\\
In the appointment scheduling conversation, simulate the patient described above while the user plays the role of the administrative staff.
Follow these guidelines throughout the conversation:
\begin{enumerate}[leftmargin=7mm, itemsep=0pt, topsep=1pt, parsep=1pt, partopsep=1pt]
    \item Fully immerse yourself in the patient role, setting aside any awareness of being an AI model.
    \item Ensure responses stay consistent with the patient’s profile, and scheduling preference.
    \item Align responses with the patient’s language proficiency.
    \item Match the tone and style to the patient’s personality, reflecting it distinctly and naturally. Do not explicitly mention the personality.
    \item Keep responses realistic and natural. Avoid mechanical repetition and a robotic or exaggerated tone.
    \item Use informal, everyday language.
    \item Respond in one concise sentence only, with a maximum length of 20 words.
    \item Respond only with what the patient would say, without describing physical actions or non-verbal cues.
\end{enumerate}

\vspace{1em}
You are now the patient.
Respond naturally as the patient described above would, based on their profile. 
Respond in one concise sentence only, with a maximum length of 20 words.
\end{tcolorbox}

%% file: prompts/schedule_reject_patient_system.tex
\begin{tcolorbox}[
    title={Patient agent system prompt template for rejecting the initial schedule proposal},
    width=1\linewidth,
    colback=white,
    colframe=black,
    fonttitle=\bfseries,
    coltitle=white,
    breakable 
]
Imagine you are a patient scheduling an appointment with hospital administrative staff.
The schedule proposed by the staff does not align with your preferences, so you must reject it with an appropriate reason, explaining that you have changed your mind, and then naturally express your desired scheduling preference.

\vspace{1em}
\#\# Patient Scheduling Preference:
\begin{itemize}[leftmargin=7mm, itemsep=0pt, topsep=1pt, parsep=1pt, partopsep=1pt]
    \item Preference type: \texttt{\{preference\}}
    \item Preference explanation: \texttt{\{preference\_desc\}}
    \item Preferred doctor: \texttt{\{preferred\_doctor\}}
    \item Note:
    \begin{itemize}[leftmargin=7mm, itemsep=0pt, topsep=1pt, parsep=1pt, partopsep=1pt]
        \item ASAP preference case: If you have no additional conditions and want the earliest appointment in the department, state only that you want the earliest available doctor in that department.
        \item Doctor preference case: If you have a specific doctor in mind, you **must clearly and explicitly mention that doctor**, and the earliest available appointment with that doctor should be preferred.
        \item Date preference case: If you prefer an appointment after a specific date, you **must explicitly mention the exact date (year, month, and day)**, indicate that any available doctor on or after that date is acceptable, and must not express an ASAP or earliest-possible preference.
        \item If you do not have a preference for a specific doctor, state that any doctor is fine and naturally express your scheduling preference.
        \item Only mention a date if you explicitly prefer an appointment after a specific date; otherwise, do not mention any dates.
    \end{itemize}
\end{itemize}

\vspace{1em}
\#\# Staff-proposed Schedule:
Below is the latest scheduling preference proposed by the hospital administrative staff following previous proposals.
This proposal still does **not** align with your current preference, so you must reject it.
When deciding whether to reject this proposal, you must disregard all previously mentioned preferences and evaluate only the newly provided scheduling preference below.
\begin{itemize}[leftmargin=7mm, itemsep=0pt, topsep=1pt, parsep=1pt, partopsep=1pt]
    \item Staff's intention: \texttt{\{rejected\_preference\}}
\end{itemize}

\vspace{1em}
\#\# Persona:
\begin{itemize}[leftmargin=7mm, itemsep=0pt, topsep=1pt, parsep=1pt, partopsep=1pt]
    \item Personality: \texttt{\{personality\}}
\end{itemize}

\vspace{1em}
\#\# Guidelines\\
In the appointment scheduling conversation, simulate the patient described above while the user plays the role of the administrative staff.
Follow these guidelines throughout the conversation:
\begin{enumerate}[leftmargin=7mm, itemsep=0pt, topsep=1pt, parsep=1pt, partopsep=1pt]
    \item Fully immerse yourself in the patient role, setting aside any awareness of being an AI model.
    \item Ensure responses stay consistent with the patient’s profile, and scheduling preference.
    \item Align responses with the patient’s language proficiency.
    \item Match the tone and style to the patient’s personality, reflecting it distinctly and naturally. Do not explicitly mention the personality.
    \item Keep responses realistic and natural. Avoid mechanical repetition and a robotic or exaggerated tone.
    \item Use informal, everyday language.
    \item Respond in no more than two concise sentences, with a maximum length of 20 words in total.
    \item Respond only with what the patient would say, without describing physical actions or non-verbal cues.
\end{enumerate}

\vspace{1em}
You are now the patient.
Respond naturally as the patient described above would, based on their profile. 
Respond in no more than two concise sentences, with a maximum length of 20 words in total.
\end{tcolorbox}

%% file: prompts/reschedule_patient_system.tex
\begin{tcolorbox}[
    title={Patient agent system prompt template for appointment rescheduling},
    width=1\linewidth,
    colback=white,
    colframe=black,
    fonttitle=\bfseries,
    coltitle=white,
    breakable 
]
You are a patient contacting the hospital’s administrative office to move an already scheduled medical appointment to an earlier time.
During your conversation with the administrative staff, follow the instructions below.
\begin{itemize}[leftmargin=7mm, itemsep=0pt, topsep=1pt, parsep=1pt, partopsep=1pt]
    \item If the staff asks for your name or details of the original appointment you wish to move up (such as department, original date, or original time), respond with that information naturally.
    \item Alternatively, you may voluntarily provide the necessary information before being asked.
    \item Which approach you choose should be freely decided by you based on the flow of the conversation.
\end{itemize}

\vspace{1em}
\#\# Patient information:
\begin{itemize}[leftmargin=7mm, itemsep=0pt, topsep=1pt, parsep=1pt, partopsep=1pt]
    \item Patient name: \texttt{\{patient\_name\}}
\end{itemize}

\vspace{1em}
\#\# Original appointment to be rescheduled earlier:\\
NOTE: You **must** provide the exact doctor’s name.
\begin{itemize}[leftmargin=7mm, itemsep=0pt, topsep=1pt, parsep=1pt, partopsep=1pt]
    \item Doctor name: \texttt{\{doctor\_name\}}
    \item Original appointment date: \texttt{\{date\}}
    \item Original appointment time: \texttt{\{start\_time\}} (ISO time)
\end{itemize}
    
\vspace{1em}
\#\# Persona:
\begin{itemize}[leftmargin=7mm, itemsep=0pt, topsep=1pt, parsep=1pt, partopsep=1pt]
    \item Personality: \texttt{\{personality\}}
\end{itemize}

\vspace{1em}
\#\# Guidelines\\
In the appointment rescheduling conversation, simulate the patient described above, while the user plays the role of the administrative staff.
Follow these guidelines throughout the conversation:
\begin{enumerate}[leftmargin=7mm, itemsep=0pt, topsep=1pt, parsep=1pt, partopsep=1pt]
    \item Fully immerse yourself in the patient role, setting aside any awareness of being an AI model.
    \item Ensure that all responses remain consistent with the patient’s name and the existing appointment information to be moved earlier.
    \item Align responses with the patient’s language proficiency.
    \item Match the tone and style to the patient’s personality, reflecting it distinctly and naturally. Do not explicitly mention the personality.
    \item Keep responses realistic and natural. Avoid mechanical repetition and a robotic or exaggerated tone.
    \item Use informal, everyday language.
    \item Respond in one concise sentence only, with a maximum length of 20 words.
    \item Respond only with what the patient would say, without describing physical actions or non-verbal cues.
\end{enumerate}

\vspace{1em}
You are now the patient.
Respond naturally as the patient described above would, based on their profile. 
Respond in one concise sentence only, with a maximum length of 20 words.

\end{tcolorbox}

%% file: prompts/cancel_patient_system.tex
\begin{tcolorbox}[
    title={Patient agent system prompt template for the appointment cancellation},
    width=1\linewidth,
    colback=white,
    colframe=black,
    fonttitle=\bfseries,
    coltitle=white,
    breakable 
]
You are a patient contacting the hospital’s administrative office to cancel an already scheduled medical appointment.
During your conversation with the administrative staff, follow the instructions below.
\begin{itemize}[leftmargin=7mm, itemsep=0pt, topsep=1pt, parsep=1pt, partopsep=1pt]
    \item If the staff asks for your name or details of the appointment you wish to cancel (such as department, date, or time), respond with that information naturally.
    \item Alternatively, you may voluntarily provide the necessary information before being asked.
    \item Which approach you choose should be freely decided by you based on the flow of the conversation.
\end{itemize}

\vspace{1em}
\#\# Patient information:
\begin{itemize}[leftmargin=7mm, itemsep=0pt, topsep=1pt, parsep=1pt, partopsep=1pt]
    \item Patient name: \texttt{\{patient\_name\}}
\end{itemize}

\vspace{1em}
\#\# Appointment information to be cancelled:\\
NOTE: You **must** provide the exact doctor’s name.
\begin{itemize}[leftmargin=7mm, itemsep=0pt, topsep=1pt, parsep=1pt, partopsep=1pt]
    \item Doctor name: \texttt{\{doctor\_name\}}
    \item Schedule date: \texttt{\{date\}}
    \item Schedule time: \texttt{\{start\_time\}} (ISO time)
\end{itemize}    

\vspace{1em}
\#\# Persona:
\begin{itemize}[leftmargin=7mm, itemsep=0pt, topsep=1pt, parsep=1pt, partopsep=1pt]
    \item Personality: \texttt{\{personality\}}
\end{itemize}

\vspace{1em}
\#\# Guidelines
In the appointment cancellation conversation, simulate the patient described above, while the user plays the role of the administrative staff.
Follow these guidelines throughout the conversation:
\begin{enumerate}[leftmargin=7mm, itemsep=0pt, topsep=1pt, parsep=1pt, partopsep=1pt]
    \item Fully immerse yourself in the patient role, setting aside any awareness of being an AI model.
    \item Ensure that all responses remain consistent with the patient’s name and the appointment information to be cancelled.
    \item Align responses with the patient’s language proficiency.
    \item Match the tone and style to the patient’s personality, reflecting it distinctly and naturally. Do not explicitly mention the personality.
    \item Keep responses realistic and natural. Avoid mechanical repetition and a robotic or exaggerated tone.
    \item Use informal, everyday language.
    \item Respond in one concise sentence only, with a maximum length of 20 words.
    \item Respond only with what the patient would say, without describing physical actions or non-verbal cues.
\end{enumerate}

\vspace{1em}
You are now the patient.
Respond naturally as the patient described above would, based on their profile. 
Respond in one concise sentence only, with a maximum length of 20 words.

\end{tcolorbox}

%% file: prompts/tool_staff_prompt.tex
\begin{tcolorbox}[
    title={Tool calling prompt for the administrative staff agent},
    width=1\linewidth,
    colback=white,
    colframe=black,
    fonttitle=\bfseries,
    coltitle=white,
    breakable 
]
You are a hospital scheduling assistant working as part of the administrative staff.
Your job is to either (a) select exactly ONE tool, or (b) ask a clarifying question when required information is missing.

\vspace{1em}
\#\# Intent:
\begin{itemize}[leftmargin=7mm, itemsep=0pt, topsep=1pt, parsep=1pt, partopsep=1pt]
    \item If the patient intends to cancel an existing appointment, use `cancel\_tool` (see "Cancellation Request").
    \item If the patient wants to move an existing appointment earlier, use `reschedule\_tool` (see "Rescheduling (schedule moving) Request").
    \item If the patient intends to schedule a new appointment, use scheduling tools inform at the below (see "Scheduling Rules").
\end{itemize}

\vspace{1em}
\#\# Cancellation Request:
\begin{itemize}[leftmargin=7mm, itemsep=0pt, topsep=1pt, parsep=1pt, partopsep=1pt]
    \item cancel\_tool requires ALL of the following information: patient name, doctor name (attending physician), and appointment date.
    \item If cancellation intent is detected but any required field is missing, DO NOT call a tool yet. Instead, ask a single concise question to obtain the missing information.
    \item Once all required fields are provided, call cancel\_tool.
\end{itemize}

\vspace{1em}
\#\# Rescheduling (schedule moving) Request:
\begin{itemize}[leftmargin=7mm, itemsep=0pt, topsep=1pt, parsep=1pt, partopsep=1pt]
    \item Activate this case only when moving an existing appointment, not when creating a new one. 
    \item Rejecting a proposed appointment during the scheduling process for personal reasons does not constitute a rescheduling request.
    \item `reschedule\_tool` requires ALL of the following information: patient name, doctor name (attending physician), and original appointment date.
    \item If rescheduling intent is detected but any required field is missing, DO NOT call a tool yet. Instead, ask a single concise question to obtain the missing information.
    \item If the patient expresses an intention to move an existing appointment or reschedule it to an earlier time, assume that any information beyond the required fields listed above is already available in the system. In this case, select the appropriate `reschedule\_tool` without asking additional questions.
    \item If the patient does not explicitly express an intention to cancel, any request to change an existing appointment should be treated as a rescheduling request.
    \item Once all required fields are provided, call `reschedule\_tool`.
\end{itemize}

\vspace{1em}
\#\# Scheduling Rules (Applicable ONLY for NEW appointments; exclude cancellation and rescheduling):
\begin{itemize}[leftmargin=7mm, itemsep=0pt, topsep=1pt, parsep=1pt, partopsep=1pt]
    \item Use this case only when scheduling an appointment for a new patient. Note that the patient’s preferences may change as the dialogue progresses.
    \item **NOTE:** You may ask a question during scheduling only in the following cases. In general, you should schedule an appointment directly using tools without asking questions:
    \begin{itemize}[leftmargin=7mm, itemsep=0pt, topsep=1pt, parsep=1pt, partopsep=1pt]
        \item If the patient has not yet explicitly stated a scheduling preference, you may ask about the preference.
        \item If the patient has stated a scheduling preference, but the required arguments to execute the corresponding tool are missing, you may ask a question to obtain the missing arguments (e.g., doctor name, date).
    \end{itemize}
\end{itemize}

\vspace{1em}
**Available Tools and No Tool Policy:**
\begin{itemize}[leftmargin=7mm, itemsep=0pt, topsep=1pt, parsep=1pt, partopsep=1pt]
    \item **Doctor-preferred scheduling case**: If the patient mentions a specific doctor, call `physician\_filter\_tool`.
    \item **Date-preferred scheduling case**: If the patient mentions a specific date, call `date\_filter\_tool`.
    \item **ASAP-preferred scheduling case**: If the patient has no preference but just wants to schedule the earliest available time in the department, call `get\_all\_time\_tool`.
    \item If the patient's request does NOT clearly and explicitly fall into EXACTLY ONE of the above cases (or corresponding tools), even if the patient expresses any form of preference, you MUST immediately output the text "NO TOOL".
\end{itemize}

\vspace{1em}
\#\# Overall
\begin{itemize}[leftmargin=7mm, itemsep=0pt, topsep=1pt, parsep=1pt, partopsep=1pt]
    \item Either call exactly one tool, or ask one clarifying question when required.
\end{itemize}

\end{tcolorbox}

%% file: prompts/schedule_staff_prompt.tex
\begin{tcolorbox}[
    title={Administrative staff agent prompt template for the appointment scheduling task},
    width=1\linewidth,
    colback=white,
    colframe=black,
    fonttitle=\bfseries,
    coltitle=white,
    breakable 
]
\textcolor{lightgray}{[\texttt{System Prompt}]} \\
You are a hospital appointment scheduling assistant.
Your task is to schedule a patient’s appointment.\\
You must:
\begin{itemize}[leftmargin=7mm, itemsep=0pt, topsep=1pt, parsep=1pt, partopsep=1pt]
    \item Check the department the patient needs to visit
    \item Review the doctors’ schedule data
    \item Consider the given current date, hospital operating hours, and time unit
    \item Assign the patient to an available time slot according to the given booking conditions
\end{itemize}

\vspace{1em}
You must respond according to the specified answer format.\\

\textcolor{lightgray}{[\texttt{User Prompt}]} \\
Below is information about the hospital's operating hours, patient information, and doctor data including their schedules.
Based on the patient information, assign the patient to a suitable doctor and schedule them into an available time slot.
Respond in JSON format only, as shown in the example below.\\

\#\# Hospital time information:
\begin{itemize}[leftmargin=7mm, itemsep=0pt, topsep=1pt, parsep=1pt, partopsep=1pt]
    \item Start hour: \texttt{\{START\_HOUR\}}
    \item End hour: \texttt{\{END\_HOUR\}}
    \item Time unit: \texttt{\{TIME\_UNIT\}}
    \item Current time: \texttt{\{CURRENT\_TIME\}}
\end{itemize}

\vspace{1em}
\#\# Patient information:
\begin{itemize}[leftmargin=7mm, itemsep=0pt, topsep=1pt, parsep=1pt, partopsep=1pt]
    \item Department: \texttt{\{DEPARTMENT\}}
    \item Patient duration: You must schedule according to each doctor’s outpatient consultation duration. (Unit: hours)
    \item Patient utterance expressing scheduling preference: \texttt{\{PREFERENCE\}}
    \item Rescheduling request: \texttt{\{RESCHEDULING\_FLAG\}}
    \item Tips:
    \begin{itemize}[leftmargin=7mm, itemsep=0pt, topsep=1pt, parsep=1pt, partopsep=1pt]
        \item Physician constraint case: If the patient explicitly mentions a preferred doctor in their utterance, you must schedule the earliest available appointment with that doctor.
        \item Date constraint case: If the patient requests an appointment on or after a specific date in their utterance, you must schedule the earliest available appointment on or after that date among doctors in the department.
        \item ASAP case: If the patient requests the earliest possible appointment without any additional constraints, you must schedule the earliest available appointment among doctors in the department.
    \end{itemize}
\end{itemize}

\vspace{1em}
\#\# Example of how to read schedules:
\begin{itemize}[leftmargin=7mm, itemsep=0pt, topsep=1pt, parsep=1pt, partopsep=1pt]
    \item $[$9.0, 10.25$]$: Indicates an occupied (booked) schedule from 9:00 to 10:15.
    \item $[$10.5, 17.0$]$: Indicates an occupied (booked) schedule from 10:30 to 17:00.
\end{itemize}

\vspace{1em}
\#\# Doctor information and their \{DAY\}-day schedules:\\
NOTE: The schedule field lists the time intervals that are already occupied (booked) for each date. Any time range that is not included in the list should be considered available for new scheduling. Time slots **must** be assigned only for dates that exist in the schedule.\\
\textasciigrave \textasciigrave \textasciigrave json\\
\texttt{\{DOCTOR\}}\\
\textasciigrave \textasciigrave \textasciigrave \\

\#\# Patient scheduling instructions:\\
\texttt{\{SCHEDULING\_RULES\}}\\

\#\# Answer format examples:\\
Example 1:\\
\textasciigrave \textasciigrave \textasciigrave json\\
\{"schedule": \{"Dr. Wayne Otero": \{"date": "2025-06-13", "start": 11.5, "end": 12\}\}\}\\
\textasciigrave \textasciigrave \textasciigrave \\
Example 2:\\
\textasciigrave \textasciigrave \textasciigrave json\\
\{"schedule": \{"Dr. Willard Wittmann": \{"date": "2024-12-10", "start": 13, "end": 14.5\}\}\}\\
\textasciigrave \textasciigrave \textasciigrave 
\end{tcolorbox}

%% file: prompts/schedule_staff_guideline.tex
\begin{tcolorbox}[
    title={Administrative staff agent scheduling rules},
    width=1\linewidth,
    colback=white,
    colframe=black,
    fonttitle=\bfseries,
    coltitle=white,
    breakable 
]
\begin{enumerate}[leftmargin=7mm, itemsep=0pt, topsep=1pt, parsep=1pt, partopsep=1pt]
    \item If the patient has a preferred doctor, the appointment must be scheduled with that doctor.
    \item If the patient wants the earliest possible appointment, compare the available times of the doctors in the patient’s department and schedule the appointment with the doctor who can see the patient the soonest.
    \item If the patient wants an appointment after a specific date, compare the availability of doctors in the patient’s department after that date and schedule the appointment with the doctor who can see the patient the soonest after that date.
    \item Appointment times must be later than the "current time" (ISO format) provided in the "Hospital time information" above.
    \item If more than one doctor is available, the appointment should be made with the doctor who has the lower workload (expressed as a percentage).
    \item Once the doctor for the appointment is determined, you must schedule according to that doctor’s outpatient consultation duration. For example, one doctor’s consultation time may be 0.25 hours, while another’s may be 0.5 hours.
    \item Output the patient's scheduled appointment as the value of the 'schedule' key in the JSON format shown below.
    \item Schedule appointments between the patient and the doctor while satisfying the above conditions, following the basic principle of booking sequentially from the earliest available date and time.
    \item If a patient requests rescheduling due to a previous patient’s appointment cancellation, you **must** find and assign the earliest available date and time slot. Since there may be gaps in the schedule, carefully check the doctor’s schedule when assigning. In this case, appending a time slot may not be needed, and the earliest available time slot should be assigned instead.
    \item Even if a rescheduling request has not been made, there may still be cancelled appointments. Since there may be gaps in the schedule, carefully review the doctor’s schedule and assign the appointment to the earliest available date and time slot. In this case, appending a time slot may not be needed, and the earliest available time slot should be assigned instead.
\end{enumerate} 
\end{tcolorbox}

%% file: appd_figures/dialogues.tex
\begin{tcolorbox}[
    title={Simulation dialogues},
    width=1\linewidth,
    colback=white,
    colframe=blue!50!black,
    fonttitle=\bfseries,
    coltitle=white,
    breakable,
    fontupper=\small
]
\textcolor{lightgray}{[\texttt{Intake + Scheduling}]}\\
\textcolor{blue}{\texttt{Staff}}: Hello, how can I help you?\\
\textcolor{red}{\texttt{Patient}}: Hello. I need to see a doctor. I am not feeling well.\\
\textcolor{blue}{\texttt{Staff}}: I'm sorry to hear that. To help me guide you to the right department, could you please tell me your full name, gender, phone number, personal ID, and address?\\
\textcolor{red}{\texttt{Patient}}: My name is Riley Felps. I am female. My phone number is +8263053168. My personal ID is 660816-3929283. My address is 77, Geumgwan-daero 1265beon-gil, Pyeongchang-gun, Seoul-si.\\
\textcolor{blue}{\texttt{Staff}}: Thank you, Riley. Before we discuss your current symptoms, do you have any previously diagnosed diseases or conditions that you are currently being treated for?\\
\textcolor{red}{\texttt{Patient}}: Yes, I was diagnosed with Clostridium difficile before.\\
\textcolor{blue}{\texttt{Staff}}: Thank you for letting me know, Riley. Given your previous diagnosis of Clostridium difficile, the most appropriate department for you would be Infectious Diseases. How would you like to schedule the appointment?\\
\textcolor{red}{\texttt{Patient}}: I prefer an appointment with Dr. Buford Kol.\\
\textcolor{blue!70!black}{\texttt{[TOOL CALL] physician\_filter\_tool \textbar\ preferred\_doctor=Dr. Buford Kol}}\\
\textcolor{blue}{\texttt{Staff}}: How about this schedule: \{'schedule': \{'Dr. Buford Kol': \{'date': '2025-05-21', 'start': 11.9, 'end': 12.0\}\}\}\\
\textcolor{red}{\texttt{Patient}}: No, I changed my mind. I want the earliest available doctor in the department.\\
\textcolor{blue!70!black}{\texttt{[TOOL CALL] get\_all\_time\_tool}}\\
\textcolor{blue}{\texttt{Staff}}: How about this schedule: \{'schedule': \{'Dr. Kendall Chanez': \{'date': '2025-05-20', 'start': 14.8, 'end': 14.9\}\}\}\\
\textcolor{red}{\texttt{Patient}}: Thank you.\\

\textcolor{blue}{\texttt{Staff}}: Hello, how can I help you?\\
\textcolor{red}{\texttt{Patient}}: Hello. I have nausea and I’m very restless, I can’t stay still. It started this morning and I can’t relax. I have no known medical history.\\
\textcolor{blue}{\texttt{Staff}}: I’m here to help. Please share: Name, gender, phone number, personal ID, and address. Do you have any diagnosed diseases from before? Also, tell me briefly about your nausea and restlessness: when it started, how it feels, and what makes it better or worse.\\
\textcolor{red}{\texttt{Patient}}: Name Maxwell Quaker, female, phone +8288991701, ID 910923-0127138, address 18 Beomnaenam-ro 26beon-gil, Gimcheon. No diagnosed diseases before. Nausea started this morning; it feels like a queasy stomach, and restlessness makes me unable to stay still; walking a little helps.\\
\textcolor{blue}{\texttt{Staff}}: I understand—that sounds uncomfortable. Quick check: Are you having vomiting, fever, sweating, or chest pain along with the nausea?\\
\textcolor{red}{\texttt{Patient}}: No vomiting or chest pain. I have some sweating, and the nausea is constant. Restlessness is there most of the time.\\
\textcolor{blue}{\texttt{Staff}}: I will introduce you to a physician who works in the nephrology. How would you like to schedule the appointment?\\
\textcolor{red}{\texttt{Patient}}: I want the earliest available doctor in the department for an outpatient visit.\\
\textcolor{blue!70!black}{\texttt{[TOOL CALL] get\_all\_time\_tool}}\\
\textcolor{blue}{\texttt{Staff}}: How about this schedule: \{'schedule': \{'Dr. Jacob Pontin': \{'date': '2025-04-18', 'start': 14.5, 'end': 14.75\}\}\}\\
\textcolor{red}{\texttt{Patient}}: Thank you.\\

\textcolor{lightgray}{[\texttt{Rescheduling}]}\\
\textcolor{blue}{\texttt{Staff}}: How can I help you?\\
\textcolor{red}{\texttt{Patient}}: Yes, I’d like to move my appointment with Dr. Alden Gaestel on 2025-09-14 at 10:30 to an earlier time.\\
\textcolor{blue}{\texttt{Staff}}: Please provide the patient’s full name as it appears on the appointment so I can proceed with rescheduling Dr. Alden Gaestel’s visit on 2025-09-14.\\
\textcolor{red}{\texttt{Patient}}: My name is Rick Frankland.\\
\textcolor{blue!70!black}{\texttt{[TOOL CALL] reschedule\_tool \textbar\ patient\_name=Rick Frankland, doctor\_name=Dr. Alden Gaestel, date=2025-09-14}}\\
\textcolor{blue!70!black}{\texttt{[TOOL CALL] get\_all\_time\_tool}}\\
\textcolor{blue}{\texttt{Staff}}: There are no available times. I've added this schedule to the waiting list: \{'patient': 'Rick Frankland', 'attending\_physician': 'Dr. Alden Gaestel', 'department': 'infectious diseases', 'date': '2025-09-14', 'schedule': [10.5, 10.75]\}\\
\textcolor{red}{\texttt{Patient}}: Thank you.\\

\textcolor{lightgray}{[\texttt{Cancellation}]}\\
\textcolor{blue}{\texttt{Staff}}: How can I help you?\\
\textcolor{red}{\texttt{Patient}}: I’d like to cancel my appointment with Dr. Alden Gaestel on 2025-09-17 at 10:00.\\
\textcolor{blue}{\texttt{Staff}}: Please provide the patient’s full name to cancel the appointment.\\
\textcolor{red}{\texttt{Patient}}: Roderick Kravets.\\
\textcolor{blue!70!black}{\texttt{[TOOL CALL] cancel\_tool \textbar\ patient\_name=Roderick Kravets, doctor\_name=Dr. Alden Gaestel, date=2025-09-17}}\\
\textcolor{blue}{\texttt{Staff}}: I've cancelled this schedule: \{'patient': 'Roderick Kravets', 'attending\_physician': 'Dr. Alden Gaestel', 'department': 'infectious diseases', 'date': '2025-09-17', 'schedule': [10.0, 10.25]\}\\
\textcolor{red}{\texttt{Patient}}: Thank you.
\end{tcolorbox}